\documentclass{article} %

\newcommand{\finalcopy}{\iclrfinalcopy}

\usepackage[dvipsnames,svgnames,x11names]{xcolor}
\usepackage{tikz}
\usetikzlibrary{arrows.meta,shapes,calc,matrix,fit,positioning,through,backgrounds,decorations.markings,fadings}
\usepackage{pgfplots}
\usepackage{pgfplotstable}
\pgfplotsset{compat=1.9}
\usepackage{xstring}

\usepgfplotslibrary{external}

\IfBeginWith*{\jobname}{fig/extern/}{\finalcopy}{}

\tikzstyle{every picture}+=[
	remember picture,
	every text node part/.style={align=center},
	every matrix/.append style={ampersand replacement=\&},
]
\tikzstyle{tight} = [inner sep=0pt,outer sep=0pt]
\tikzstyle{node}  = [draw,circle,tight,minimum size=12pt,anchor=center]
\tikzstyle{op}    = [draw,circle,tight]
\tikzstyle{dot}   = [fill,draw,circle,inner sep=1pt,outer sep=0]
\tikzstyle{pt}    = [fill,draw,circle,inner sep=1.5pt,outer sep=.2pt]
\tikzstyle{box}   = [draw,rectangle,inner sep=3pt]
\tikzstyle{high}  = [black!60]
\tikzstyle{group} = [high,box,opacity=.5]
\tikzstyle{dim1}  = [fill opacity=.3,text opacity=1]
\tikzstyle{dim2}  = [fill opacity=.5,text opacity=1]
\tikzstyle{dim3}  = [fill opacity=.7,text opacity=1]
\tikzstyle{rectc} = [tight,transform shape]
\tikzstyle{rect}  = [rectc,anchor=south west]

\newcommand{\leg}[1]{\addlegendentry{#1}}

\tikzset{every mark/.append style={solid}}
\pgfplotsset{%
	grid=both, width=\columnwidth, try min ticks=5,
	every axis/.append style={font=\small},
	every axis plot/.append style={thick,mark=none,mark size=1.8,tension=0.18},
	legend cell align=left, legend style={fill opacity=0.8},
	nodes near coords math/.style={
		nodes near coords={\pgfmathprintnumber[assume math mode=true]{\pgfplotspointmeta}},
	},
}

\pgfplotsset{
	dash/.style={mark=o,dashed,opacity=0.6},
	dott/.style={mark=o,dotted,opacity=0.6},
	nolim/.style={enlargelimits=false},
	plain/.style={every axis plot/.append style={},nolim,grid=none},
}

\pgfdeclarelayer{bg4}
\pgfdeclarelayer{bg3}
\pgfdeclarelayer{bg2}
\pgfdeclarelayer{bg1}
\pgfdeclarelayer{fg1}
\pgfdeclarelayer{fg2}
\pgfdeclarelayer{fg3}
\pgfdeclarelayer{fg4}
\pgfsetlayers{bg4,bg3,bg2,bg1,main,fg1,fg2,fg3,fg4}

\makeatletter

\tikzdeclarecoordinatesystem{rel}{%
    \tikzset{cs/.cd,x=0pt,y=0pt,#1}%
    \pgfpointlineattime{(\tikz@cs@x)}%
       {\pgfpointanchor{\tikz@pp@name{\tikz@cs@node}}{south west}}%
       {\pgfpointanchor{\tikz@pp@name{\tikz@cs@node}}{south east}}%
    \edef\tikz@cs@x{\the\pgf@x}%
    \pgfpointlineattime{(\tikz@cs@y)}%
       {\pgfpointanchor{\tikz@pp@name{\tikz@cs@node}}{south west}}%
       {\pgfpointanchor{\tikz@pp@name{\tikz@cs@node}}{north west}}%
    \pgfpoint{\tikz@cs@x}{\pgf@y}%
  }

\def\tikz@parse@relcs#1(#2:#3,#4){%
\tikz@parse@coordinatesystem#1(rel cs:name={#2},x={#3},y={#4})%
}
\let\tikz@parse@relcs@polar@saved=\tikz@parse@polar
\def\tikz@parse@polar#1(#2:#3){%
  \pgfutil@in@{,}{#3}%
  \ifpgfutil@in@%
    \let\@next\tikz@parse@relcs%
  \else%
    \let\@next\tikz@parse@relcs@polar@saved%
  \fi%
  \@next#1(#2:#3)%
}

\makeatother

\usepackage{iclr2024_conference}

\usepackage{times}
\usepackage{amsmath}
\usepackage{amsfonts}
\usepackage{amssymb}
\usepackage{url}
\usepackage{graphicx}
\usepackage{epigraph}
\usepackage{bm}
\usepackage{bbm}

\usepackage{capt-of}
\usepackage{caption}
\usepackage{subcaption}
\usepackage{colortbl}
\usepackage{booktabs}
\usepackage{multirow}

\usepackage{pifont}
\usepackage{adjustbox}    
\usepackage{float}
\usepackage{cuted}
\usepackage{tabu}
\usepackage[shortlabels]{enumitem}
\usepackage[toc,page,header]{appendix}
\usepackage{minitoc}

\usepackage{xspace}
\usepackage[pagebackref,breaklinks,colorlinks,citecolor=blue, linkcolor=blue]{hyperref}
\usepackage[capitalize]{cleveref}
\usepackage{mathrsfs}

\def\eqref#1{equation~\ref{#1}}

\def\1{\bm{1}}

\def\vs{{\bm{s}}}

\DeclareMathAlphabet{\mathsfit}{\encodingdefault}{\sfdefault}{m}{sl}
\SetMathAlphabet{\mathsfit}{bold}{\encodingdefault}{\sfdefault}{bx}{n}

\newcommand{\alert}[1]{{\color{red}{#1}}}

\newcommand{\eq}[1]{(\ref{eq:#1})}

\newcommand{\Th}[1]{\textsc{#1}}
\newcommand{\mr}[2]{\multirow{#1}{*}{#2}}
\newcommand{\mc}[2]{\multicolumn{#1}{c}{#2}}
\newcommand{\tb}[1]{\textbf{#1}}

\newcommand{\red}[1]{{\textcolor{red}{#1}}}

\newcommand{\gray}[1]{{\textcolor{gray}{#1}}}

\newcommand{\citeme}[1]{\alert{[X]}}
\newcommand{\refme}[1]{\alert{(X)}}

\newcommand{\fig}[2][1]{\includegraphics[width=#1\linewidth]{fig/#2}}

\newcommand{\tran}{^\top}

\newcommand{\ind}{\mathbbm{1}}

\newcommand{\real}{\mathbb{R}}

\newcommand{\softmax}{\operatorname{softmax}}

\newcommand{\defn}{\mathrel{:=}}

\newcommand{\wt}[1]{\widetilde{#1}}

\newcommand{\cF}{\mathcal{F}}

\newcommand{\cI}{\mathcal{I}}
\newcommand{\cJ}{\mathcal{J}}

\newcommand{\vT}{\mathbf{T}}

\newcommand{\vX}{\mathbf{X}}

\makeatletter
\newcommand*\bdot{\mathpalette\bdot@{.7}}
\newcommand*\bdot@[2]{\mathbin{\vcenter{\hbox{\scalebox{#2}{$\m@th#1\bullet$}}}}}
\makeatother

\makeatletter
\DeclareRobustCommand\onedot{\futurelet\@let@token\@onedot}
\def\@onedot{\ifx\@let@token.\else.\null\fi\xspace}

\def\eg{\emph{e.g}\onedot} 
\def\ie{\emph{i.e}\onedot} 
 
\def\etc{\emph{etc}\onedot} \def\vs{\emph{vs}\onedot}

\makeatother

\newcommand{\Ours}{DORA\xspace}
\newcommand{\ours}{\textsc{DoRA}\xspace}
\newcommand{\our}{\textsc{DoRA}}

\newcommand{\WT}{WTours\xspace}
\newcommand{\Wt}{WT\xspace}
\newcommand{\Wtv}{WT$_\text{Venice}$\xspace}
\newcommand{\Wta}{WT$_\text{all}$\xspace}

\newcommand{\cls}{\textsc{[cls]}\xspace}

\newcommand{\sk}{\operatorname{SK}}

\definecolor{Gray}{gray}{0.85}
\definecolor{LightCyan}{rgb}{0.88,1,1}
\definecolor{ForestGreen}{RGB}{34,139,34}

\newcommand{\gp}[1]{\color{ForestGreen}{#1}} 

\newcommand{\cmark}{\ding{51}}%
\newcommand{\xmark}{\ding{55}}%

\newcolumntype{a}{>{\columncolor{Gray}}c}

\iclrfinalcopy

\title{Is ImageNet worth 1 video? Learning strong image encoders from 1 long unlabelled video}

\author{Shashanka Venkataramanan \\
Inria, Univ Rennes, CNRS, IRISA
\And
Mamshad Nayeem Rizve\\
University of Central Florida
\And
Jo\~ao Carreira \\
Google DeepMind 
\And
Yuki M. Asano\thanks{Equal last authors. Order determined randomly}\\
University of Amsterdam
\And
Yannis Avrithis$^\ast$\\
Institute of Advanced Research \\ on Artificial Intelligence (IARAI)
}
\makeatletter
\renewcommand\paragraph{\@startsection{paragraph}{4}{\z@}{0.6ex}{-1em}{\normalfont\normalsize\bfseries}}
\makeatother
\newcommand{\midsepremove}{\aboverulesep = 0.2mm \belowrulesep = 0.3mm}
\midsepremove

\newcommand{\yuki}[1]{}
\newcommand{\iavr}[1]{}
\newcommand{\shashank}[1]{}
\newcommand{\joao}[1]{}

\usepackage[symbol]{footmisc}

\begin{document}

\maketitle
\doparttoc %
\faketableofcontents 

\begin{abstract}
Self-supervised learning has unlocked the potential of scaling up pretraining to billions of images, since annotation is unnecessary. But are we making the best use of data? How more economical can we be? In this work, we attempt to answer this question by making two contributions. First, we investigate first-person videos and introduce a ``Walking Tours'' dataset. These videos are high-resolution, hours-long, captured in a single uninterrupted take, depicting a large number of objects and actions with natural scene transitions. They are unlabeled and uncurated, thus realistic for self-supervision and comparable with human learning. 

Second, we introduce a novel self-supervised image pretraining method tailored for learning from continuous videos. Existing methods typically adapt image-based pretraining approaches to incorporate more frames. Instead, we advocate a ``tracking to learn to recognize'' approach. Our method called~\ours, leads to attention maps that \tb{D}isc\tb{o}ver and t\tb{RA}ck objects over time in an end-to-end manner, using transformer cross-attention.
We derive multiple views from the tracks and use them in a classical self-supervised distillation loss. Using our novel approach, a single Walking Tours video remarkably becomes a strong competitor to ImageNet for several image and video downstream tasks. Dataset and code can be found at  \url{https://shashankvkt.github.io/dora}.
\end{abstract}

\section{Introduction}\label{sec:intro}
\vspace{-16pt}
\epigraphwidth=22em
\epigraph{\textit{(To the question ``Have you read all the books in here?'')} No, only four of them. But I read those very, very carefully.}{Jacques Derrida}
Learning from large scale datasets has been at the core of great progress. In particular, the field of self-supervised learning has allowed pretraining of neural networks to scale beyond the size of labelled datasets. By avoiding costly annotation, strong performance has been demonstrated by increasing the training dataset sizes into billions of images.

But how well are those images really used? At a rate of one image per second, a dataset of 1B images would take 317 years to watch. Yet, humans develop functioning visual systems \textit{much faster}.\footnote{Humans develop face recognition~\citep{de2001recognition} and color sensitivity~\citep{adams1987evaluation} in three months, depth perception in five months~\citep{campos1978emergence} and visual acuity in six months~\citep{sokol1978measurement}.}
Besides potential genetic visual priors in humans, one stark difference is the \textit{type} of data. Humans 
observe their visual surroundings in one continuous stream, only interrupted by sleep. Indeed, learning visual representations of images from videos is not new. However, previous works have 
found significant gaps in performance to 
image-pretrained 
models. 
They
have mostly used object-centric videos scraped from the internet, and adapted image-based pretraining methods to 
use
different frames as an extra form of data augmentation~\citep{gordon2020watching,parthasarathy2022self}.

In this work, we investigate two directions. First, in the direction of \emph{data}, we introduce a new dataset of open-source first-person videos, recorded for the purpose of virtual ``walking tours'', inspired by~\citep{wiles2022compressed}. These videos have several advantages. Not only are the individual frames dense in semantic categories -- much more so than movies, as we
analyze
-- but these videos also directly represent the viewpoint of a human, contain few or no shot cuts nor special effects and are long (1-3h). Another benefit is their transparency: indeed, one can watch the whole dataset in one setting.
The dataset we create contains 10 Walking Tours (\Wt) videos 
with CC-BY license.

Second, in the direction of the \emph{method}, we develop a new self-supervised image-pretraining method that is uniquely suited for learning from natural, non-object-centric videos. Our approach is inspired by observing toddlers first learn to track objects and animals, then to 
recognize and differentiate them~\citep{bomba1983nature, quinn1993evidence,spelke2007core}. Our method, called \ours, is an end-to-end training approach that ``tracks to learn to recognize'': given a 
video
clip, objects in an initial frame are implicitly \tb{D}isc\tb{o}vered and t\tb{RA}cked across time. The tracked objects are incentivized to be 
diverse
by introducing a Sinkhorn-Knopp clustering of patch embeddings; the tracked instances are used as a learning signal for a classical multi-view SSL loss.

Surprisingly, contrary to previous works, we find that our novel method obtains ImageNet-level performances by training on \emph{a single \Wt video}, as evidenced by performances on segmentation and object detection downstream tasks. While humorously intentioned, Derrida's quote rings true to this finding and our results give some hope for alternative directions in SSL that depart from blind dataset scaling towards more efficient and smarter use of existing video data.

To summarize, our key contributions in this work are as follows:
\begin{enumerate}[itemsep=1pt, parsep=0pt, topsep=0pt]
    \item  We introduce a new dataset of 10 WT videos, with single-video and mixed-video splits. The latter is conveniently equal in size to ImageNet. We analyze their usefulness compared to existing video and image datasets.
    \item We propose a new end-to-end self-supervised visual pretraining method called \ours. It builds upon DINO but is tailored to promote tracking of multiple objects across frames. We use it to learn strong image encoders and trace the source of its improvements through extensive ablations.
    \item We obtain strong performance on ADE20k segmentation and MS COCO detection, outperforming ImageNet-pretrained DINO, while instead pretraining on a single long video.
\end{enumerate}

\section{Related Work}
\label{sec:related}

\begin{figure}
\centering
\fig{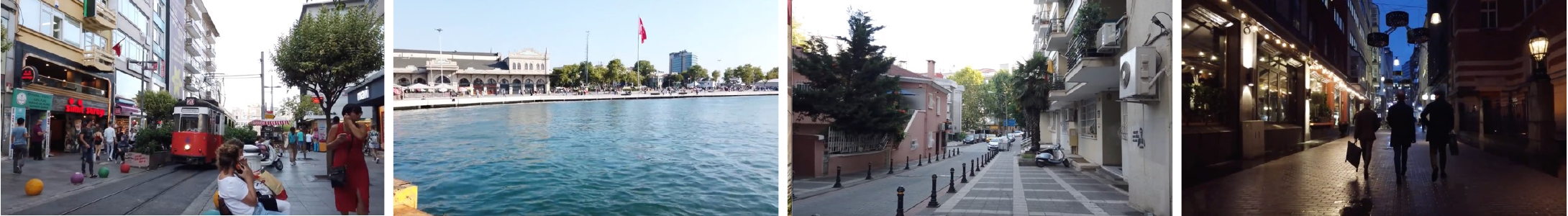}
\vspace{-1em}
\caption{\emph{Examples of frames from the Walking Tours dataset}, containing hours-long, continuous egocentric 4K videos from urban scenes in different cities, under CC-BY license. There are a large number of objects and actions in a variety and natural transition of places, \eg residential area, park, market, waterfront, \etc, with natural transition of lighting conditions and object augmentations.}
\vspace{-3pt}
\label{fig:wtour-frames}
\end{figure}

Self-supervised learning of image encoders from video data is a very active area of research. Video, and more generally temporal streams, have long been theorized to be ideal signals for unsupervised learning~\citep{wiskott2002slow}. In computer vision, early methods have been very diverse and included pretext tasks such as egomotion prediction~\citep{agrawal2015learning, jayaraman2015learning}, active recognition~\citep{jayaraman2016look}, pose estimation~\citep{Chakraborty2017Learning}, unsupervised object discovery~\citep{croitoru2017unsupervised}, dense prediction~\citep{pathak2017learning, li2019joint}, optical flow~\citep{mahendran1, xiong2021self},  frame order prediction~\citep{misra2016shuffle}, view-point matching~\citep{sermanet2018time, pirk2020online} or learning visual correspondences~\citep{wang2019learning}.

More recently, there have been considerable advances in self-supervised learning using ImageNet, with the main theme being extracting multiple augmentations of an image~\citep{chen2020simple,caron2021emerging} and training models to pull them together/apart.  These methods have since percolated to learning from video frames~\citep{gordon2020watching, parthasarathy2022self,tschannen2020self, wang2015unsupervised, orhan2020self}.
Similar to this work, TimeTuning~\citep{salehi2023time} leverages the passage of time in videos by not treating it as simple augmentations. However, in contrast to our work, it requires an already image-pretrained backbone. VITO~\citep{parthasarathy2022self}
improves performance relative to ImageNet, by using VideoNet, a large YouTube dataset of 10s videos from a similar class distribution and the same number of examples as ImageNet. In this paper, we show that it is possible to obtain strong results from a \textit{single} long video, with a very different visual distribution compared to ImageNet / VideoNet.

\section{Walking Tours Dataset}
\label{sec:dataset}

\subsection{Dataset collection and properties}
\label{sec:wt}

We collect from YouTube a new dataset of urban scenes called ``Walking Tours'' (\WT, or \Wt) comprising 10 egocentric videos of a person walking in different cities in Europe and Asia. The cities include Amsterdam, Bangkok, Chiang Mai, Istanbul, Kuala Lampur, Singapore, Stockholm, Venice, and Zurich. We also include a video from a Wildlife safari. Examples are shown in~\autoref{fig:wtour-frames}. These videos are captured in 4K resolution (3840 $\times$ 2160 pixels) at 60 frames-per-second and are under Creative Commons License (CC-BY). The minimum video duration is 59 minutes (Wildlife safari), the maximum is 2 hours 55 minutes (Bangkok) and the average is 1 hour 38 minutes. 
Such videos are particularly interesting for visual learning because of the following properties:
\begin{enumerate}[itemsep=1pt, parsep=0pt, topsep=0pt]
	\item \emph{Large number of objects and actions}. Each frame or clip taken from a video depicts several objects and actions, \eg walking, riding a bike, sitting, drinking \etc.
	\item \emph{Natural transition in lighting conditions}. In some videos, the lighting gradually transitions from bright (late afternoon) to dim (dusk) then to dark (post sunset).
	\item \emph{Natural transition in scenes}. The videos depict transitions between places, \eg from city center to market place to residential areas to parks to water fronts \etc.
	\item \emph{Natural object augmentations}. Continuous variation \eg of pose, deformation, viewpoint, perspective distortion, relative object position, occlusion, background clutter.
\end{enumerate}

The abundance of information within these videos, encompassing a multitude of objects and complex scenes, presents a formidable challenge for manual annotation or curation, making it appropriate for unsupervised pretraining. To the best of our knowledge, we are the first to propose an egocentric video dataset for pretraining and evaluate it on a wealth of downstream tasks.

\begin{table}
\centering
\scriptsize
\setlength{\tabcolsep}{3pt}
\begin{tabular}{lccccccccc} \toprule
\mr{2}{\Th{Dataset}}                      & \mr{2}{\Th{Domain}} & \mr{2}{\Th{Ego}} & \mr{2}{\Th{Pre}} & \mr{2}{\Th{Bal}} & \mr{2}{\Th{Annot}} & \Th{Avg. Dur} & \Th{Dur}  & \mr{2}{\Th{\#Videos}} & \Th{Frame}         \\
                                          &                     &                  &                  &                  &                    & \Th{(sec)}    & \Th{(hr)} &                       & \Th{resolution}    \\ \midrule
\mc{10}{\emph{Diverse Pretraining}}                                                                                                                                                                                    \\ \midrule
Kinetics-400~\citep{kay2017kinetics}      & Actions             & \red{\xmark}     & \gp{\cmark}      & \red{\cmark}     & \red{\tb{Class}}   & 10.2          & 851       & 400                   & 340 $\times$ 255   \\
WebVid-2M~\citep{bain2021frozen}          & Open                & \red{\xmark}     & \gp{\cmark}      & \gp{\xmark}      & \red{\tb{Weak}}    & 18            & 13k       & --                    & 320 $\times$ 240   \\
HowTo100M~\citep{miech2019howto100m}      & Instructions        & \red{\xmark}     & \gp{\cmark}      & \gp{\xmark}      & \red{\tb{Weak}}    & 4             & 135k      & --                    &  --                  \\ \midrule
\mc{10}{\emph{Egocentric}}                                                                                                                                                                                             \\ \midrule
Epic-Kitchens~\citep{Damen2022RESCALING}   & Cooking             & \gp{\cmark}      & \red{\xmark}     & \gp{\xmark}      & \red{\tb{Loc.}}    & 510           & 100       & 37                    & 1920 $\times$ 1080 \\
Ego-4D~\citep{grauman2022ego4d}           & Daily               & \gp{\cmark}      & \red{\xmark}     & \gp{\xmark}      & \red{\tb{Loc.}}    & 1446          & 120       & 931                   & 1920 $\times$ 1080 \\
Meccano~\citep{ragusa_MECCANO_2023}       & Industry            & \gp{\cmark}      & \red{\xmark}     & \gp{\xmark}      & \red{\tb{Loc.}}    & 1247          & 849       & 20                    & 1920 $\times$ 1080 \\
Assembly-101~\citep{sener2022assembly101} & Assembly            & \gp{\cmark}      & \red{\xmark}     & \gp{\xmark}      & \red{\tb{Loc.}}    & 426           & 167       & 362                   & 1920 $\times$ 1080 \\ \midrule
\mc{10}{\emph{ImageNet-aligned}}                                                                                                                                                                                       \\ \midrule
R2V2~\citep{gordon2020watching}           & ImageNet            & \red{\xmark}     & \gp{\cmark}      & \red{\cmark}     & \red{\tb{Class}}   & --            & --        & --                      & 467 $\times$ 280   \\
VideoNet~\citep{parthasarathy2022self}    & ImageNet            & \red{\xmark}     & \gp{\cmark}      & \red{\cmark}     & \red{\tb{Class}}   & 10            & 3055      & --                    &                    \\ \midrule \rowcolor{LightCyan}
Walking Tours (ours)                             & Urban               & \gp{\cmark}      & \gp{\cmark}      & \gp{\xmark}      & \gp{\tb{None}}     & 5880          & 23        & 10                    & 3840 $\times$ 2160 \\ \bottomrule
\end{tabular}
\caption{\emph{Walking Tours \vs existing video datasets}. \Th{Ego}: egocentric; \Th{Pre}: used for pretraining; \Th{Bal}: class balance control; \Th{Annot}: annotation type. Weak: associated data per clip (text or other modality); Class: class label per frame or clip; Loc: localization per frame (\eg bounding box, segmentation, mask 3D pose). \Th{Avg. Dur}: average duration per video; \Th{Dur}: total duration.}
\label{tab:sota-dataset}
\end{table}

\subsection{Comparison with other video datasets}
\label{sec:data-comp}

In \autoref{tab:sota-dataset}, we compare \WT with existing video datasets. In summary, self-supervised pretraining on videos has been mostly limited to short, low-resolution datasets that rely on weak annotation in the form of video-text pairs~\citep{bain2021frozen, miech2019howto100m} or are curated, \eg their class balance is controlled, even if their annotation is unused~\citep{kay2017kinetics}. \emph{ImageNet-aligned} datasets~\citep{gordon2020watching, parthasarathy2022self} contain short videos that are (semi-automatically) curated and annotated with the same distribution and classes as ImageNet. \emph{Egocentric} video datasets~\citep{Damen2022RESCALING, grauman2022ego4d, sener2022assembly101} 
 have long, high-quality videos, but are the result of significant manual work. In this paper we aim to learn from videos publicly available online.

\WT videos are continuous, longer and higher-resolution than even other egocentric datasets. Using object detectors, we find that the average number of object classes is close to that of ImageNet and there is a high number of objects per frame, making \WT appropriate for representation learning. \WT is not curated and does not rely on search terms. It is \emph{data-first} and more open-ended, thus well suited for the self-supervised setting. It is scalable since it requires no human labeling effort and more videos can be easily downloaded or even made. We are inspired by a 10k walking tours videos created by~\cite{wiles2022compressed}, which however is not publicly released and not studied for self-supervised learning. A more detailed discussion is given in \autoref{app:data-comp}.

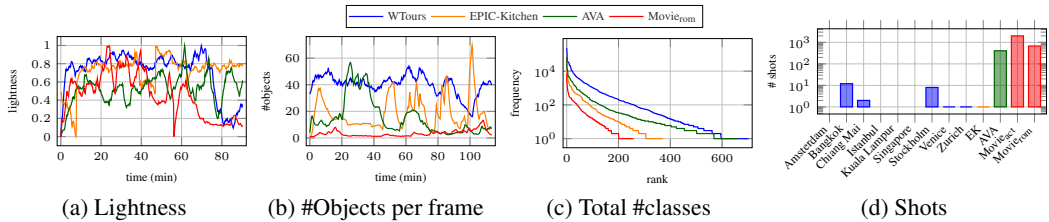
\begin{figure}
\small
\centering
\setlength{\tabcolsep}{0pt}
\newcommand{\fs}{.76}
\newcommand{\fw}{.35}
\newcommand{\fh}{.25}
\begin{tabular}{cccc}
\mc{4}{\scalebox{\fs}{\tiny\ref*{stat}}} \\[-6pt]
\scalebox{\fs}{
\begin{tikzpicture}
\begin{axis}[%
	width=\fw\linewidth,
	height=\fh\linewidth,
	font=\tiny,
	xlabel={time (min)},
	ylabel={lightness},
	enlarge x limits=.02,
	legend columns=-1,
	legend to name=stat,
	legend style={inner sep=1pt,outer sep=0pt},
	every axis plot/.append style={semithick},
]
	\pgfplotstableread{fig/stat/lightness_sub.txt}{\data}
	\addplot[blue] table[x=time,y=WT] \data; \leg{\WT};
	\addplot[orange] table[x=time,y=EK] \data; \leg{EPIC-Kitchen};
	\addplot[DarkGreen] table[x=time,y=AVA] \data; \leg{AVA};
	\addplot[red] table[x=time,y=rom] \data; \leg{Movie$_\text{rom}$};
\end{axis}
\end{tikzpicture}
}
&
\scalebox{\fs}{
\begin{tikzpicture}
\begin{axis}[%
	width=\fw\linewidth,
	height=\fh\linewidth,
	font=\tiny,
	xlabel={time (min)},
	ylabel={\#objects},
	enlarge x limits=.02,
	every axis plot/.append style={semithick},
]
	\pgfplotstableread{fig/stat/objects_sub.txt}{\data}
	\addplot[blue] table[x=time,y=WT] \data;
	\addplot[orange] table[x=time,y=EK] \data;
	\addplot[DarkGreen] table[x=time,y=AVA] \data;
	\addplot[red] table[x=time,y=rom] \data;
\end{axis}
\end{tikzpicture}
}
&
\scalebox{\fs}{
\begin{tikzpicture}
\begin{semilogyaxis}[%
	width=\fw\linewidth,
	height=\fh\linewidth,
	font=\tiny,
	xlabel={rank},
	ylabel={frequency},
	enlarge x limits=.02,
	every axis plot/.append style={semithick},
]
	\pgfplotstableread{fig/stat/classes_WT.txt}{\wt}
	\pgfplotstableread{fig/stat/classes_EK.txt}{\ek}
	\pgfplotstableread{fig/stat/classes_AVA.txt}{\ava}
	\pgfplotstableread{fig/stat/classes_rom.txt}{\rom}
	\addplot[blue] table[x=rank,y=count] \wt;
	\addplot[orange] table[x=rank,y=count] \ek;
	\addplot[DarkGreen] table[x=rank,y=count] \ava;
	\addplot[red] table[x=rank,y=count] \rom;
\end{semilogyaxis}
\end{tikzpicture}
}
&
\scalebox{\fs}{
\begin{tikzpicture}
\begin{semilogyaxis}[%
	width=.4\linewidth,
	height=.22\linewidth,
	font=\tiny,
	ylabel={\# shots},
	every axis plot/.append style={semithick},
	ybar,
	bar shift=0,
	bar width=6pt,
	xmin=.3,xmax=13.7,
	yminorgrids=false,
	try min ticks=13,
	xtick={1,2,3,4,5,6,7,8,9,10,11,12,13},
	xticklabels={Amsterdam,Bangkok,Chiang Mai,Istanbul,Kuala Lampur,Singapore,Stockholm,Venice,Zurich,EK,AVA,Movie$_\text{act}$,Movie$_\text{rom}$},
	x tick label style={rotate=45,anchor=east},
	bar/.style={fill,fill opacity=.5},
]
	\pgfplotstableread{fig/stat/shots_WT.txt}{\wt}
	\pgfplotstableread{fig/stat/shots_EK.txt}{\ek}
	\pgfplotstableread{fig/stat/shots_AVA.txt}{\ava}
	\pgfplotstableread{fig/stat/shots_mov.txt}{\mov}
	\addplot[blue,bar] table[x=video,y=shots] \wt;
	\addplot[orange,bar] table[x=video,y=shots] \ek;
	\addplot[DarkGreen,bar] table[x=video,y=shots] \ava;
	\addplot[red,bar] table[x=video,y=shots] \mov;
\end{semilogyaxis}
\end{tikzpicture}
}
\\
(a) Lightness & (b) \#Objects per frame & (c) Total \#classes & (d) Shots
\end{tabular}
\caption{\emph{Dataset analysis} of a \WT video compared with videos from Epic-Kitchens~\citep{Damen2022RESCALING}, AVA~\citep{gu2018ava} and two entire movies, concatenated or cropped to match the duration of the \WT video. (a) Lightness \vs time. (b) Number of objects per frame \vs time. (c) Frequency of classes in entire video. (d) Number of shots. Objects detected by Detic~\citep{zhou2022detecting}, trained on ImageNet-21k.
}
\label{fig:dataset-stats}
\vspace{-3pt}
\end{figure}

\subsection{Dataset analysis}
\label{sec:data-analysis}

In \autoref{fig:dataset-stats}, we analyse the properties of a single \WT video compared with videos of the same length from two other datasets, as well as two movie videos. In summary, our findings are as follows. From \autoref{fig:dataset-stats}(a), \Wt may exhibit gradual shifts in lightness, transitioning from bright to dim to dark, while Epic-Kitchens
and AVA videos exhibit random brightness fluctuations. Lightness variations are not well expored in self-supervised pretraining. From \autoref{fig:dataset-stats}(b,c), unique classes appear more frequently and there are more unique objects per frame in \WT than in the other datasets. This makes \WT semantically richer.
From \autoref{fig:dataset-stats}(d), \WT and Epic-Kitchens videos contain only one or two shots per entire video on average, while the other datasets contain hundreds. In \autoref{sec:ablation} and in \autoref{app:ablation}, we show that \WT significantly outperforms movies in downstream tasks, which  is partially attributed to the absence of cuts. More detailed discussion of dataset analysis is given in \autoref{app:data-analysis}.

\section{Attention-based multi-object tracking}
\label{sec:method}

Our goal is to build robust representations by leveraging the rich information in video frames. Standard SSL frameworks \citep{chen2020simple, caron2020unsupervised} often assume correspondences between different views. This is true whether using dense \citep{zhou2021ibot} or global representations by pooling \citep{caron2021emerging}{}. While it is relatively straightforward to establish correspondences in images, it becomes more challenging when dealing with temporal deformations, requiring some form of object tracking~\citep{salehi2023time}. In videos with a large field of view or ego-motion, obtaining correspondences becomes even more difficult.

\begin{figure}
\centering
\fig[1]{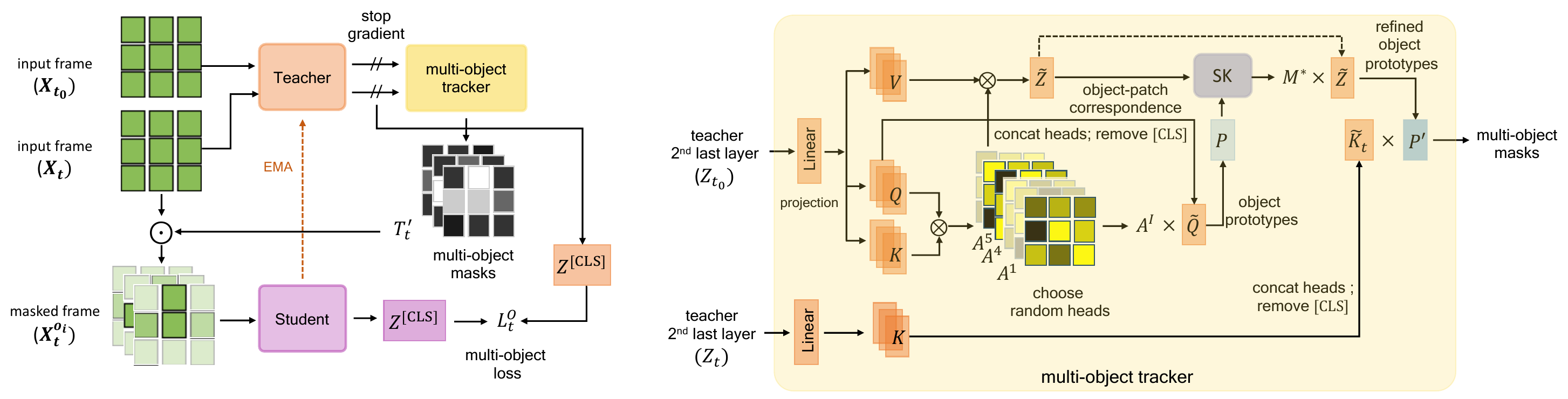}
\vspace{-3pt}
\caption{\emph{\ours, our self-supervised image pretraining method from video}. (Left) From an input frame $\vX_{t_0}$, the output of the second-last layer of the teacher model is used by a multi-object tracker to generate cross-attention maps $T'_t$ with frame $\vX_t$. We use those to mask $\vX_t$~\eq{mask}, feed it to the student model and apply a \emph{distillation loss} $L_t^{\text{O}}$ between \cls token embeddings~\eq{obj-loss}. (Right) In the tracker, we obtain the query $Q$, key $K$ and output $Z$ embeddings. From the multi-head attention maps $A^i$~\eq{self}, we draw a subset $\cI$ of $k$ heads and form object prototypes $P$ by pooling over patch queries $\tilde{Q}$~\eq{proto-1}. We refine them into $P'$ to \emph{discover distinct objects}, using Sinkhorn-Knopp (SK) to establish correspondences $M^*$ between $P$ and patch embeddings $\tilde{Z}$~\eq{sk} and pooling over $\tilde{Z}$~\eq{proto-2}. We then \emph{track the objects} over frames $\vX_t$ by cross-attention $T'_t$ with patch key embeddings $\tilde{K}_t$~\eq{cross-2}.}
\label{fig:model}

\end{figure}

\paragraph{High-level idea}
\iavr{Add diagram.}
We introduce \ours, based on multi-object \tb{D}isc\tb{o}very and t\tb{RA}cking. As shown in \autoref{fig:model}, it leverages the attention from the \cls token of distinct heads in a vision transformer to identify and consistently track multiple objects within a given frame across temporal sequences. On these, a teacher-student distillation loss is then applied. 
Importantly, we do not use any off-the-shelf object tracker or optical flow network. This keeps our pipeline simple and does not require any additional data or training. It also ensures that the learned representation is robust.

\paragraph{Preliminaries}

We are given a 
video clip consisting of $T$ frames $\vX_t \in \real^{h \times w \times c}$ for $t \in \{1,\dots,T\}$, where $h \times w$ is the spatial resolution and $c$ is the number of channels. Each frame is split into $n = hw/p^2$ non-overlapping patches of resolution $p \times p$. The patches are linearly projected into embeddings of dimension $d$ and a \cls token embedding is prepended.
This representation is input to a \emph{transformer encoder}~\citep{dosovitskiy2020image}. The \emph{output embeddings} are $Z_t = g_\theta(\vX_t) \in \real^{(n+1) \times d}$, where mapping $g_\theta$ includes the tokenizer and encoder, while $\theta$ denotes its learnable parameters. Given an embedding $Z \in \real^{(n+1) \times d}$, we write $Z = [Z^\cls; \tilde{Z}]$, where $Z^\cls \in \real^{1 \times d}$ is the \cls token embedding and $\tilde{Z} \in \real^{n \times d}$ are the \emph{patch embeddings}.

Following DINO~\citep{caron2021emerging}, there is a student network with parameters $\theta$ and a teacher network with identical architecture and parameters $\theta'$ obtained as the \emph{exponential moving average} (EMA) of $\theta$ according to $\theta' \gets \alpha \theta' + (1-\alpha) \theta$. The encoder is followed by a \emph{head} that includes an MLP and a scaled softmax, such that the output token embeddings can be interpreted as probabilities. We denote by $f_\theta$ the mapping that includes the tokenizer, encoder and head.

\paragraph{Discovering objects with multi-head attention}

Starting at a first frame $\vX_{t_0}$, we obtain the query and key embeddings $Q, K \in \real^{(n+1) \times d}$ from the last transformer layer of the teacher network\footnote{For simplicity, we drop $t_0$ from the notation.}. According to \emph{multi-head} attention, these embeddings are partitioned as $Q = [Q^1, \dots, Q^h]$, $K = [K^1, \dots, K^h]$, where $Q^i, K^i \in \real^{(n+1) \times d/h}$ for $i=1,\dots,h$ and $h$ is the number of heads. For each head $i$, the \emph{self-attention} matrix $A^i \in \real^{(n+1) \times (n+1)}$ is based on the dot-product similarity between the query and key embeddings:
\begin{equation}
	A^i \defn \softmax \left( Q^i (K^i)\tran / \sqrt{d} \right) \in \real^{(n+1) \times (n+1)}.
\label{eq:self}
\end{equation}
Given an attention matrix $A \in \real^{(n+1) \times (n+1)}$, let $A^\cls \defn [a_{1,2}, \dots, a_{1,n}] \in \real^{1 \times n}$ be the \emph{\cls-attention} vector between the \cls and patch embeddings, where $a_{i,j}$ is the element $(i,j)$ of $A$. We draw at random a subset $\cI \defn \{i_1,\dots,i_k\}$ of $k<h$ heads and collect their \cls-attention vectors into $A^\cI \defn [(A^{i_1})^\cls; \dots; (A^{i_k})^\cls] \in \real^{k \times n}$. Intuitively, as expressed in rows of matrix $A^\cI$, the different heads attend to different \emph{objects} in the frame~\citep{caron2021emerging}.

\begin{figure}
\small
\centering
\setlength{\tabcolsep}{1pt}
\newcommand{\sz}{.16}
\newcommand{\cen}{.07\linewidth}
\newcommand{\node}[2]{\tikz[remember picture]{\node[tight](#1){#2};}}
\newcommand{\ellipse}[5][thick,red]{\tikz[remember picture,overlay]{%
	\node[draw,#1,shift={(#2:#3)},circle through={(#2:.5,0)},xscale=#4,yscale=#5] {};
}}
\begin{tabular}{cccccc}
&
$t=1$ &
$t=2$ &
$t=3$ &
$t=4$ &
$t=5$ \\
\raisebox{\cen}{$\vX_t$} &
\fig[\sz]{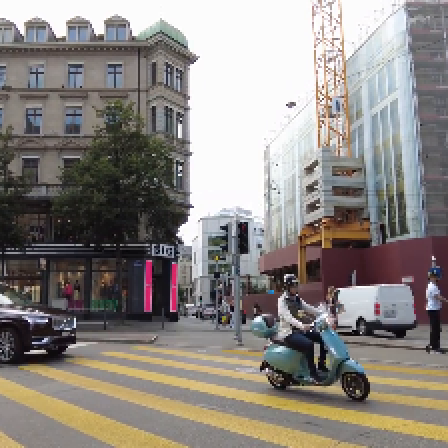} &
\fig[\sz]{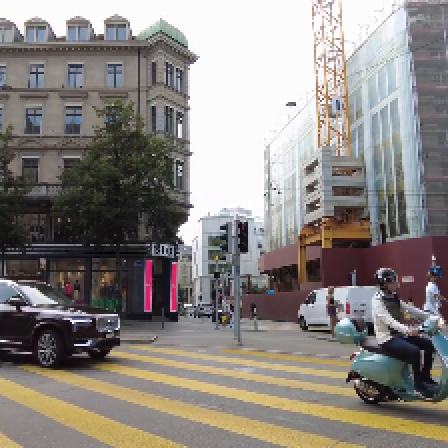} &
\fig[\sz]{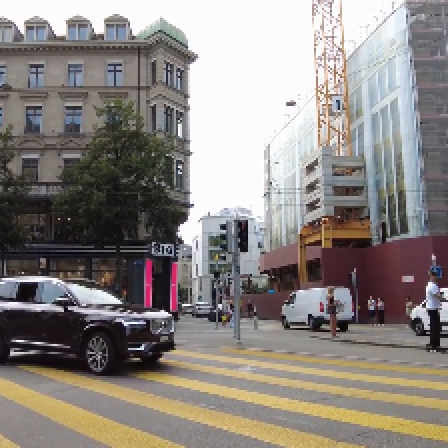} &
\fig[\sz]{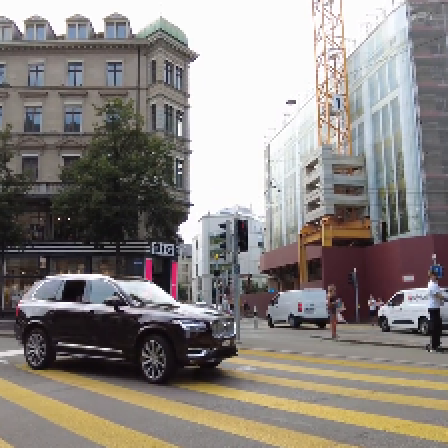} &
\fig[\sz]{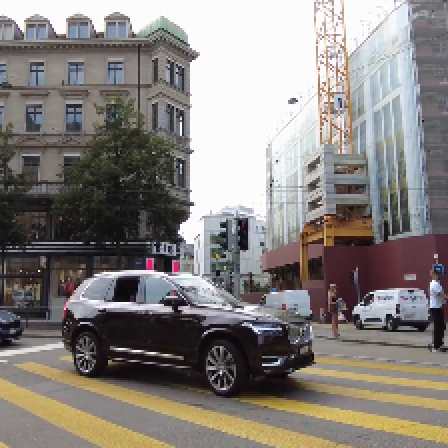} \\[-1pt]
\raisebox{\cen}{$T_t$} &
\node{b}{\fig[\sz]{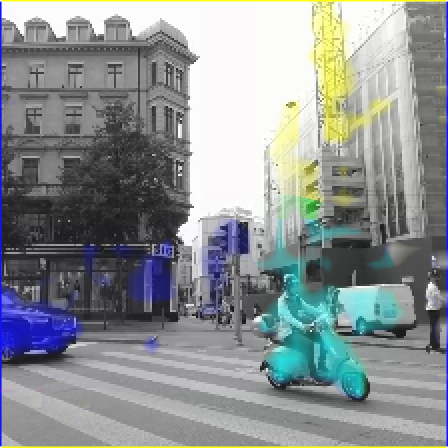}} \ellipse{b}{.7,.3}{.4}{.5} &
\node{b}{\fig[\sz]{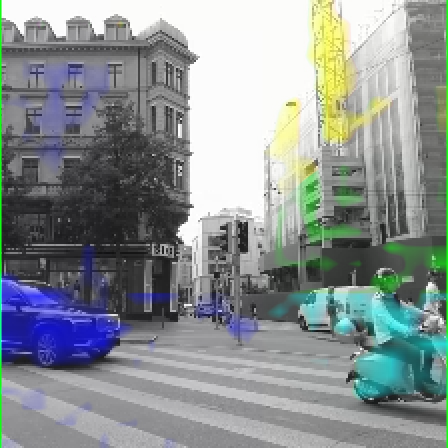}} \ellipse{b}{.7,.8}{.3}{.4} &
\fig[\sz]{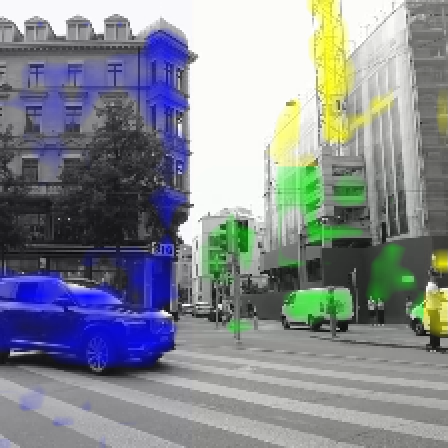} &
\fig[\sz]{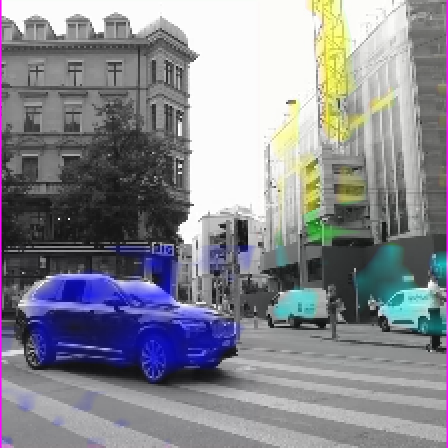} &
\fig[\sz]{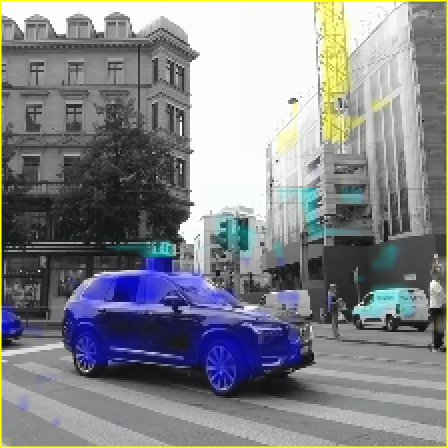} \\[-1pt]
\raisebox{\cen}{$T'_t$} &
\fig[\sz]{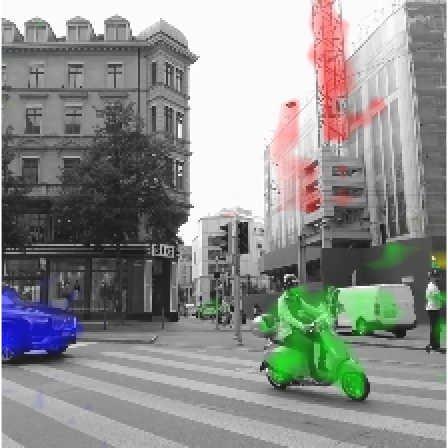} &
\fig[\sz]{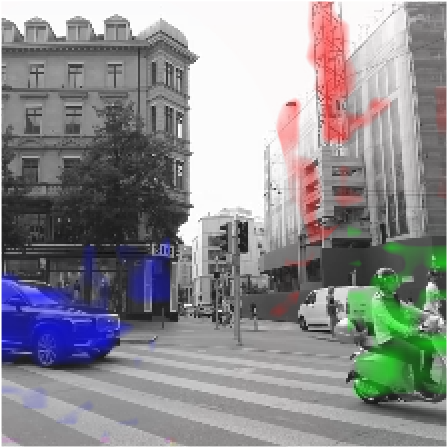} &
\fig[\sz]{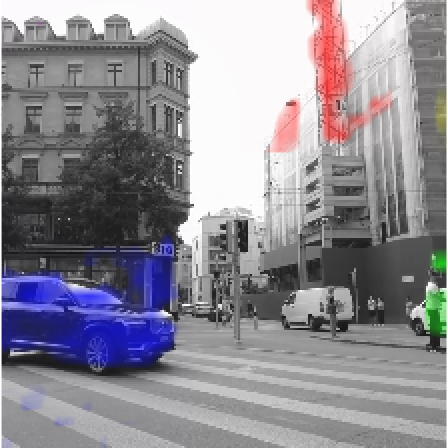} &
\fig[\sz]{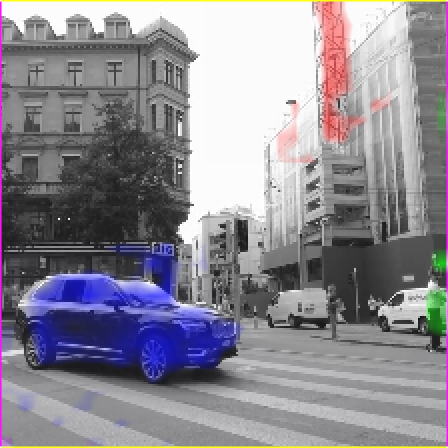} &
\fig[\sz]{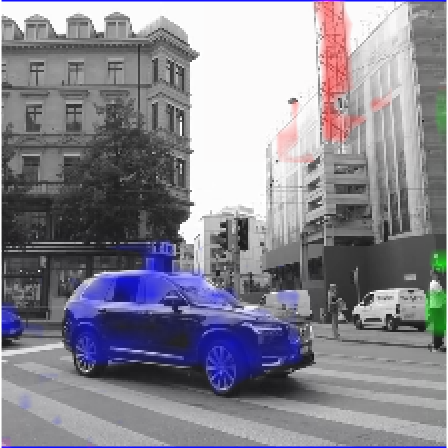} \\
\end{tabular}
\caption{For each input frame $t$ of a video clip (top), cross-attention map $T_t \in \real^{k \times n}$~\eq{cross-1} (middle) and refined cross-attention map $T'_t \in \real^{k \times n}$~\eq{cross-2} (bottom), using Sinkhorn-Knopp algorithm. For each object, one row of $T_t$ or $T'_t$ is reshaped as $h/p \times w/p$ and upsampled to an $h \times w$ \emph{attention map} overlaid on the input frame for $k=3$ objects encoded in blue, red and green channel. Mixed colors yellow and cyan for $T_t$ (middle, in red circle) indicate spatial overlap of two objects, while $T'_t$ (bottom) yields three well separated objects shown in primary colors blue, red and green.
\label{fig:tracking}}

\end{figure}

To represent the $k$ objects in the embedding space, we use matrix $A^\cI \in \real^{k \times n}$ to form linear combinations of patch embeddings $\tilde{Q} \in \real^{n \times d}$, obtaining \emph{object prototypes}
\begin{equation}
	P \defn A^\cI \tilde{Q} \in \real^{k \times d}.
\label{eq:proto-1}
\end{equation}
This can be seen as the representation of $k$ different \cls tokens in the full embedding space, capturing $k$ objects at frame $t_0$. Then, given the key embeddings $K_t \in \real^{(n+1) \times d}$ at another frame $t$, we could track the objects by \emph{cross-attention}
\begin{equation}
	T_t \defn \softmax \left( P \tilde{K}_t\tran / \sqrt{d} \right) \in \real^{k \times n},
\label{eq:cross-1}
\end{equation}
where $\tilde{K}_t \in \real^{n \times d}$. Unfortunately, we observe in \autoref{fig:tracking} that the $k$ attention maps obtained this way are spatially overlapping, meaning that each attention map is not delineating a single object.

\paragraph{Establishing object-patch correspondences}

To discover spatially distinct objects, we propose to establish correspondences between prototypes and patch tokens. Let $Z = g_{\theta'}(\vX_{t_0}) \in \real^{(n+1) \times d}$ be the output embeddings of the teacher network, still at frame $t_0$. We seek a correspondence between the rows of $P \in \real^{k \times d}$ and $\tilde{Z} \in \real^{n \times d}$, where $\tilde{Z}$ are the patch token embeddings.

The goal is to find a \emph{transport plan} $M \in \real^{k \times n}$ that minimizes the expected pairwise cost $C \defn -P \tilde{Z}\tran \in \real^{k \times n}$ between prototypes and patches, while incorporating an entropic regularizer with coefficient $\epsilon$. Matrix $M$ is non-negative with row-wise sum $1/k$ and column-wise sum $1/n$, representing a joint probability over $P$ and  $\tilde{Z}$ with uniform marginals. The minimal solution $M^*$ is unique and can be found by forming the matrix $e^{-C/\epsilon}$ and then applying the Sinkhorn-Knopp (SK) algorithm~\citep{cuturi2013sinkhorn}, %
\ie, iteratively normalizing its rows and columns:
\begin{equation}
	M^* = \sk \left( \exp \left( P \tilde{Z}\tran / \epsilon \right) \right) \in \real^{k \times n},
\label{eq:sk}
\end{equation}
Observe the similarity with~\eq{self} and~\eq{cross-1}, where scaling is by $\sqrt{d}$ rather than $\epsilon$, $\exp$ is included in softmax and normalization is on rows only rather than iterative. Then, similarly with~\eq{proto-1}, we use the optimal transport plan $M^* \in \real^{k \times n}$ to form linear combinations of patch embeddings $\tilde{Z} \in \real^{n \times d}$, obtaining the \emph{refined object prototypes}
\begin{equation}
	P' = M^* \tilde{Z} \in \real^{k \times d}.
\label{eq:proto-2}  
\end{equation}
Now, given the key embeddings $K_t \in \real^{(n+1) \times d}$ at another frame $t$, we track the objects by the \emph{refined cross-attention}, similarly with~\eq{cross-1}:
\begin{equation}
	T'_t \defn \softmax \left( P' \tilde{K}_t\tran / \sqrt{d} \right) \in \real^{k \times n},
\label{eq:cross-2}
\end{equation}
where $\tilde{K}_t \in \real^{n \times d}$. Indeed, \autoref{fig:tracking} confirms that each of the $k$ resulting attention maps is associated with a spatially distinct object, thanks to the established correspondences.

In contrast to previous works that use SK in the context of self-supervised learning to force an equi-partitioning of images to cluster labels~\citep{asano2020self, caron2020unsupervised, oquab2023dinov2}, we rather use optimal transport to re-balance \emph{spatial} correspondences to different objects.

\paragraph{Multi-object masking}

We use the cross-attention~\eq{cross-2} to mask the input video clip for the student network, such that each masked clip can be considered as a \emph{multi-object crop}. This crop plays a similar role with local crops in DINO~\citep{caron2021emerging}, but it has arbitrary shape and tracks an object over video frames. In particular, given an input frame $\vX \in \real^{h \times w \times c}$ with cross-attention matrix $T' \in \real^{k \times n}$~\eq{cross-2} and an object $i \in \{1,\dots,k\}$, we reshape the $i$-th row of $T'$ as $h/p \times w/p$ and upsample to a $h \times w$ attention map to match the spatial resolution of $\vX$, as shown in \autoref{fig:tracking}. We repeat along the channel dimension to form tensor $\vT^i \in \real^{h \times w \times c}$ and we mask $\vX$ as
\begin{equation}
	\vX^{o_i} \defn \vX \odot \vT^i,
\label{eq:mask}
\end{equation}
where $\odot$ is the Hadamard product. Following DINO~\citep{caron2021emerging}, given an input frame $\vX_t$, we generate two standard resolution augmented \emph{global views} $\vX_t^a, \vX_t^b$. We introduce a \emph{multi-object loss} 
$L_t^{\text{O}}$ for frame $t$, applied to the \cls token between the teacher $f_{\theta'}$ output for one global view $\vX_t^u$ and the student $f_\theta$ output for the masked version $\vX_t^{v,o_i}$ of the other view $\vX_t^v$ for $i \in \{1,\dots,k\}$, where $u,v \in V = \{a, b\}$ and $u \ne v$:
\begin{equation}
	L_t^{\text{O}} \defn \sum_{u,v \in V} \ind_{u \ne v} \sum_{i=1}^k
		f_{\theta'}(\vX_t^u)^\cls
		\log \left(f_\theta(\vX_t^{v,o_i})^\cls \right).
\label{eq:obj-loss}
\end{equation}
In addition, as detailed in \autoref{app:multi-crop}, we apply a \emph{local loss}, following \emph{multi-crop}~\citep{caron2020unsupervised}. The overall loss $L$ is the sum of the two losses, averaged over all $T$ frames.

\section{Experiments}
\label{sec:exp}

\subsection{Setup}
\label{sec:setup}

\paragraph{Tasks and methods}

We perform self-supervised pretraining on a single \Wt tour video in Venice (referred to as \Wtv) or all 10 \Wt videos (referred to as \Wta) and compare with other image and video datasets. To evaluate the quality of the learned representations, we use frozen features for classification, unsupervised object discovery and video object segmentation. We fine-tune for semantic segmentation, object detection and object tracking. We compare \ours with SoTA SSL methods~\citep{Guilherme2022solo} using our settings. We provide more details in individual sections per task. Implementation details and hyperparameters are given in \autoref{app:setup}.

\subsection{Ablations}
\label{sec:ablation}

We examine the effect of using different pretraining video dataset and different options and parameters for \ours, measuring performance of classification on ImageNet-1k~\citep{deng2009imagenet} by linear probing (LP) accuracy and unsupervised object discovery on Pascal-VOC 2012~\citep{pascal-voc-2012} by correct localization (CorLoc)~\citep{simeoni2021localizing}.

\begin{table}[tb]
\footnotesize
\centering
	\begin{subtable}[b]{0.4\linewidth}\centering
		{\centering
\scriptsize
\setlength{\tabcolsep}{2.5pt}
\begin{tabular}{llccc} \toprule
\Th{Method}              & \Th{Pretrain}      & \Th{\#Frames}  & \Th{LP}   & \Th{CorLoc} \\
                         &                    & (M)            &           &             \\ \midrule
DINO                     & Movie$_\text{rom}$ & 0.19              & 34.9      & 51.5        \\ \rowcolor{LightCyan}
\ours                    & Movie$_\text{rom}$ & 0.19              & \tb{35.3} & \tb{51.6}   \\ \midrule
DINO                     & K-400$^*$          & 0.2               & 40.7      & 52.4        \\ \rowcolor{LightCyan}
\ours                    & K-400$^*$          & 0.2               & \tb{43.0} & \tb{55.2}   \\ \midrule
DINO                     & EK$^*$             & 0.2               & 38.6      & 53.5        \\ \rowcolor{LightCyan}
\ours                    & EK$^*$             & 0.2               & \tb{41.8} & \tb{56.0}   \\ \midrule
DINO                     & \Wtv               & 0.2               & 33.8      & 51.2        \\ \rowcolor{LightCyan}
\ours                    & \Wtv               & 0.2               & \tb{44.5} & \tb{56.2}   \\ \bottomrule

\end{tabular}
\caption{\emph{Video datasets}}
\label{tab:diff-dataset}

}
	\end{subtable}
    \hfill
	\begin{subtable}[b]{0.26\linewidth}\centering
        {\centering
\scriptsize
\setlength{\tabcolsep}{2pt}
\begin{tabular}{lcccccc} \toprule
\Th{Method}  & $k$          & \Th{LP}   & \Th{CorLoc} \\ \midrule  
DINO         & \red{\xmark} & 33.8      & 51.2        \\ \midrule
\ours        & 1            & 39.9      & 53.9            \\
\ours        & 2            & 43.1      & 55.7            \\
\ours        & 3            & \tb{44.5} & \tb{56.2}   \\
\ours        & 4            & 39.2      & 53.8            \\
\ours        & 5            & 36.7      & 50.3            \\
\ours        & 6            & 35.8      & 48.8            \\
\ours        & 16           & 28.3      & 48.5        \\
\ours        & 32           & 27.1      & 46.8        \\ \bottomrule
\end{tabular}
\caption{\emph{\#Objects $k$ on \Wtv}}
\label{tab:abl-obj}
}
        \end{subtable}
    \hfill
	\begin{subtable}[b]{0.3\linewidth}\centering
		{\centering
\scriptsize
\setlength{\tabcolsep}{2pt}
\begin{tabular}{lllcccc} \toprule
\Th{Method}    & \Th{SK}      & \Th{Mask}    & \Th{LP}   & \Th{CorLoc} \\ \midrule  
DINO           & \red{\xmark} & \red{\xmark} & 33.8      & 51.2        \\ \midrule
\ours          & \red{\xmark} & Random       & 33.0      & 49.8            \\
\ours          & \red{\xmark} & Object       & 42.5      & 55.3        \\
\ours          & \gp{\cmark}  & Random       & 29.9      & 46.7         \\
\ours          & \gp{\cmark}  & Object       & \tb{44.5} & \tb{56.2}   \\ \bottomrule
\end{tabular}
\caption{\emph{SK and masking on \Wtv}}
\label{tab:abl-sk-mask}
}
	\end{subtable}
\vspace{-6pt}
\caption{\emph{Effect of parameters}. ViT-S/16 pretrained, then frozen. (a) Different pretraining video dataset, (b) Number $k$ of tracked objects. (c) Random or multi-object mask, without SK~\eq{cross-1} and with SK~\eq{cross-2}. $^*$: subset of videos with same total duration as a single \WT video. K-400: Kinetics-400, EK: Epic-Kitchens. LP: top-1 accuracy (\%) of linear probing on the validation set of ImageNet-1k. CorLoc: correct localization on validation set of Pascal-VOC 2012.}
\label{table:ablation}
\end{table}

\paragraph{Pretraining video dataset}

We study the impact of pretraining on diverse video datasets, encompassing \emph{object-centric} videos such as Kinetics-400 (K-400)~\citep{kay2017kinetics}, \emph{egocentric} videos like Epic-Kitchens (EK)~\citep{Damen2022RESCALING} and a single movie, Movie$_\text{rom}$~\citep{movie_rom}. To maintain uniformity in terms of the number of frames, we curate a subset of videos from K-400 and EK, such that their total duration is the same as a single \Wt video. In \autoref{tab:diff-dataset}, we observe that although K-400
is \emph{object-centric}, pretraining on \WT videos yields superior performance on ImageNet and Pascal-VOC 2012. Pretraining on a single movie yields is inferior to both \WT and K-400 by a large margin. This is possibly due to the presence of cuts, as studied in \autoref{app:ablation}.

\paragraph{Number of tracked objects}

We study the impact of the number $k$ of objects. Objects are discovered using attention heads, where the total number of heads is in ViT-S/16 is $h=6$. For $k>h$, we modify the MSA block as described in \autoref{app:impl}. In \autoref{tab:abl-obj}, we observe that $k=3$ works best. We hypothesize that this is a compromise between the number of objects that can be tracked and the multi-object loss~\eq{obj-loss} attempting to match small objects with the global crop.

\paragraph{Choice of masking and Sinkhorn-Knopp}

We explore the effect of using a multi-object mask~\eq{mask} \vs random block-wise~\citep{zhou2021ibot} and the effect of improving object-patch correspondence through SK in refined cross-attention~\eq{cross-2} \vs~\eq{cross-1}. In~\autoref{tab:abl-sk-mask}, we observe that a multi-object mask leads to a remarkable performance improvement even in the absence of SK. In fact, random block-wise mask undermines object-patch correspondence, making the effect of SK negative. By contrast, SK improves performance in the presence of multi-object mask.

\subsection{Comparison with State-of-the-art}
\label{sec:results}

\begin{table}
\centering
\scriptsize
\setlength{\tabcolsep}{4pt}
\begin{tabular}{lcccccc|cc|cc} \toprule
\mr{2}{\Th{Method}}                       & \mr{2}{\Th{epochs}} & \mr{2}{\Th{Pretrain}} & \mc{4}{(a) \Th{Semantic seg.}}                              & \mc{2}{(b) \Th{Object det.}}     & \mc{2}{(c) \Th{Instance seg.}}                      \\ \cmidrule{4-11}
                                          &                     &                       & mIoU        & \tiny \Th{\gp{Gain}} & Acc$_m$     & \tiny \Th{\gp{Gain}} & mAP         & \tiny \Th{\gp{Gain}} & mIoU        & \tiny \Th{\gp{Gain}} \\ \midrule
ViT-S/16                                  & 100                 & none                  & 25.1        &                & 33.3        &                & 28.6         &                & 24.3         &                \\ \midrule
iBOT\tiny~\citep{zhou2021ibot}            & 100                 & \Wtv                  &  33.9       &                & 43.3        &                &  37.6       &                & 33.0        &                \\
AttMask\tiny~\citep{kakogeorgiou2022hide} & 100                 & \Wtv                  &  33.6       &                & 42.7        &                & 36.5        &                &  32.5       &                \\
VITO\tiny~\citep{parthasarathy2022self}   & 300                 & VideoNet              & 39.4        &                & --          &                & 44.0        &                & --          &                \\
DINO\tiny~\citep{caron2021emerging}       & 100                 & IN-1k                 & 33.9        &                & 44.3        &                & 39.9        &                & 35.1        &                \\ \midrule
\gray{\ours (ours)}                              & \gray{100}          & \gray{\Wta}         & \gray{36.9} &                & \gray{48.0} &                & \gray{40.7} &                & \gray{36.3} &                \\ \midrule
DINO\tiny~\citep{caron2021emerging}       & 100                 & \Wtv                  & 32.4        &                & 43.7        &                & 37.1        &                & 32.1        &                \\ \rowcolor{LightCyan}
\ours (ours)                                    & 100                 & \Wtv                  & \tb{35.4}   & \gp{+3.0}      & \tb{45.5}   & \gp{+1.8}      & \tb{39.5}   & \gp{+2.4}      & \tb{34.7}   & \gp{+2.6}      \\ \bottomrule
\end{tabular}
\vspace{-3pt}
\caption{\emph{Semantic segmentation, object detection and instance segmentation}. ViT-S/16 pretrained, then fine-tuned. \Wtv: Walking Tours (ours), single video of \emph{Venice}; \Wta: all videos. IN-1k: ImageNet-1k. (a) Semantic segmentation: fine-tuning on ADE20k using UperNet. mIoU: mean IoU; Acc$_m$: mean-class accuracy. (b) Object detection and (c) Instance segmentation: fine-tuning on MS-COCO using Cascade RCNN. mAP: mean average precision; mIoU: mean IoU.}
\label{tab:ade-coco}
\vspace{-3pt}
\end{table}

\paragraph{Dense scene understanding}

\autoref{tab:ade-coco}(a) shows \emph{semantic segmentation} by fine-tuning on ADE20k~\citep{zhou2017scene} using UperNet~\citep{xiao2018unified}. \ours outperforms DINO by 3\% mIoU, and 1.8\% Acc$_m$. It is interesting to note that \Ours pretrained on 200k frames of a \emph{single} \WT video outperforms DINO pretrained on 1.3M images of ImageNet-1k
by 1.5\% mIoU. A more comparable setting is \ours pretrained on 1.5M frames of \Wta, which outperforms DINO pretrained on ImageNet by 3\% mIoU.
\autoref{tab:ade-coco}(b) shows \emph{object detection} and \emph{instance segmentation} by fine-tuning on MS-COCO~\citep{lin2014microsoft} using Cascade RCNN~\citep{cai2019cascade}. \ours outperforms DINO by 2.4\% mAP and 2.6\% mIoU. \ours pretrained on \Wta outperforms DINO pretrained on ImageNet by 0.8\% mIoU and 1.2\% mAP. This shows that pretraining on \WT videos significantly improves the generality of \ours to dense prediction tasks, requiring only one tenth of the total images.

\begin{table}
\centering
\scriptsize
\setlength{\tabcolsep}{2pt}
\begin{tabular}{lcccccccc|ccccccc} \toprule
\mr{2}{\Th{Method}}                     & \mr{2}{\Th{epochs}} & \mr{2}{\Th{Pretrain}} & \mc{6}{(a) \Th{Video object segmentation}}                                                   & \mc{6}{(b) \Th{Object tracking}}                                                       \\ \cmidrule{4-15}
                                        &                     &                       & $(\cJ\&\cF)_m$ &\Th{\gp{Gain}} & $\cJ_m$   &\Th{\gp{Gain}} & $\mathcal{F}_m$ &\tiny \Th{\gp{Gain}} & mAO       & \tiny \Th{\gp{Gain}} & SR$_{0.5}$ & \tiny \Th{\gp{Gain}} & SR$_{0.75}$ & \Th{\gp{Gain}} \\ \midrule
ViT \tiny ~\citep{dosovitskiy2020image}   & 100           & None                & 26.9            &               & 25.4       &               & 28.3             &               & 23.1     &                & 19.0	       &               & 3.4        &                \\ \cmidrule{2-15}
iBOT \tiny ~\citep{zhou2021ibot}        & 100                 & \Wtv                  &  57.4              &               & 56.7          &               & 58.0                &               &    41.5       &                &        47.5    &               &        16.6     &                \\ 
DINO \tiny ~\citep{caron2021emerging}   & 100                 & IN-1k                 & 59.4           &               & 57.4      &               & 61.4            &               & 46.4      &                & 54.3       &               & 24.1        &                \\ \midrule
\gray{\ours (ours)}                            & \gray{100}          & \gray{\Wta}         & \gray{57.6}               &           & \gray{55.1}          &               & \gray{60.2}                &               &      \gray{45.9}    &                &   \gray{53.4}         &               &       \gray{23.7}      &                \\ \midrule
DINO \tiny ~\citep{caron2021emerging}   & 100                 & \Wtv                  & 54.6           &               & 53.0      &               & 56.2            &               & 37.4      &                & 41.4	    &               & 13.4        &                \\ \rowcolor{LightCyan}
\ours (ours)                                  & 100                 & \Wtv                  & \tb{58.4}      & \gp{+3.8}     & \tb{56.4} & \gp{+3.4}     & \tb{60.4}       & \gp{+4.2}     & \tb{41.4} & \gp{+4.0}      & 47.2       & \gp{+5.8}     & 18.2        & \gp{+4.8}      \\ \bottomrule
\end{tabular}
\vspace{-3pt}
\caption{\emph{Video object segmentation and object tracking}. ViT-S/16 pretrained, then frozen or fine-tuned. \Wtv: Walking Tours (ours), single video from \emph{Venice}; \Wta: all videos. IN-1k: ImageNet-1k. (a) Video object segmentation: frozen features on DAVIS-2017. $\cJ_m$: mean region similarity; $\cF_m$: mean contour-based accuracy. (b) Multi-object tracking: fine-tuning on on GOT-10k. mAO: mean average overlap; SR: success rate, threshold 50\% and 75\%.}
\label{tab:vos-got}
\end{table}

\paragraph{Video understanding}

\autoref{tab:vos-got}(a) shows video object segmentation by using frozen features on DAVIS-2017~\citep{pont20172017}, which assesses the ability to segment an object over its dynamic temporal changes. \ours captures detailed temporal deformations and outperforms baseline DINO by 3.4\% $\cJ_m$ and 4.2\% $\cF_m$. Using only a \emph{single} video for pretraining, \ours achieves almost the same performance of DINO pretrained on ImageNet (56.4\% \vs 57.4\% $\cJ_m$). \autoref{tab:vos-got}(b) shows multi-object tracking by fine-tuning on GOT-10k~\citep{Huang2021} using SeqTrack~\citep{chen2023seqtrack}. GOT-10k assesses the ability to track extremely fast moving objects, objects with illumination variation and low resolution. \ours achieves significant gains between 4-6\% over DINO. 

\begin{table}
\centering
\scriptsize
\setlength{\tabcolsep}{3pt}
\begin{tabular}{lccccccc|cccc} \toprule
\mr{2}{\Th{Method}}                       & \mr{2}{\Th{epochs}} & \mr{2}{\Th{Pretrain}} & \Th{\#Frames} & \mc{4}{(a) \Th{Classification}}                                & \mc{4}{(b) \Th{Object discovery}}                          \\ \cmidrule{5-12}
                                          &                     &                       & (M)           & \Th{LP}   & \tiny \Th{\gp{Gain}}      & \Th{$k$-NN} &\tiny \Th{\gp{Gain}} & \Th{Jacc.} & \tiny \Th{\gp{Gain}} & \Th{CorLoc} & \tiny \Th{\gp{Gain}} \\ \midrule
SimCLR\tiny~\citep{chen2020simple}        & 100                 & \Wtv                  & 0.2           & 26.3      &                     & 25.9        &                & 40.4           &                &  50.2           &                \\
SwAV\tiny~\citep{caron2020unsupervised}   & 100                 & \Wtv                  & 0.2           & 28.0      &                     & 26.4        &                &  40.6          &                &  51.4           &                \\
iBOT\tiny~\citep{zhou2021ibot}            & 100                 & \Wtv                  & 0.2           & 36.8      &                     & 32.8        &                & 43.0           &                &  53.1           &                \\
AttMask\tiny~\citep{kakogeorgiou2022hide} & 100                 & \Wtv                  & 0.2           & 35.8      &                     & 31.9        &                & 43.5           &                & 54.5            &                \\
VicReg\tiny~\citep{bardes2021vicreg}      & 100                 & \Wtv                  & 0.2           & 36.5      &                     & 30.1        &                &  42.7          &                &  52.1           &                \\ \midrule
DINO\tiny~\citep{caron2021emerging}       & 100                 & \Wtv                  & 0.2           & 33.8      &                     & 29.9        &                & 43.8       &                & 51.2        &                \\ \rowcolor{LightCyan}
\ours (ours)                                    & 100                 & \Wtv                  & 0.2           & \tb{45.4} & \gp{+11.6}          & \tb{33.8}   & \gp{+3.9}      & \tb{44.0}  & \gp{+0.2}      & \tb{56.2}   & \gp{+5.0}               \\ \midrule
DINO\tiny~\citep{caron2021emerging}       & 100                 & \Wta               & 1.5           & 36.6      &                     & 31.1        &                & 42.9       &                & 55.8        &                \\ \rowcolor{LightCyan}
\ours (ours)                                    & 100                 & \Wta               & 1.5           & \tb{45.3} & \gp{+8.7}           & \tb{35.7}   & \gp{+4.6}      & \tb{44.3}  & \gp{+1.4}      & \tb{57.1}   & \gp{+1.3}      \\ \bottomrule
\end{tabular}
\vspace{-3pt}
\caption{\emph{Image classification and object discovery}. ViT-S/16 pretrained, then frozen. \Wtv: Walking Tours (ours), single video from \emph{Venice}; \Wta: all videos. (a) Classification top-1 accuracy (\%) on validation set of ImageNet-1k. LP: linear probing. (b) Unsupervised object discovery on validation set of Pascal-VOC 2012. Jacc.: Jaccard similarity; CorLoc: Correct Localization.}
\label{tab:sota-cls}
\end{table}

\paragraph{Image classification and unsupervised object discovery}

We pretrain \ours on \WT and then we keep it frozen on the downstream task, indicating the quality of the pretrained features. \autoref{tab:sota-cls}(a) shows \emph{image classification} on ImageNet-1k, measuring accuracy for linear probing and $k$-nearest neighbor. \autoref{tab:sota-cls}(b) shows \emph{unsupervised object discovery} on Pascal-VOC 2012,
using attention maps as segmentation masks to measure Jaccard similarity and CorLoc. 

On both tasks, non-contrastive methods (DINO, iBOT, VICReg) outperform contrastive methods (SimCLR, SwAV), when pretrained on a single \Wt video. Importantly, non-contrastive methods are also more efficient to train, since no negative pairs are used. Also on both tasks, \ours outperforms DINO by a large margin, \eg 11.6\% LP and 3.9\% $k$-NN on classification, when trained on a single \Wt video. Comparing \ours on \Wtv with the \Wta dataset, the improvement brought by the full dataset is small when using \ours, although it is 10 times larger.

In \autoref{app:results}, we show that \ours outperforms SoTA methods on all tasks on ImageNet-1k.

\section{Conclusions}

We have introduced a dataset of 10 walking tour videos -- first-person videos taken by people touring a city, with no cuts, high resolution and that are hours long. We show that learning from clips taken from these videos is surprisingly powerful: with an appropriately tailored self-supervised learning method for videos, we obtain representations that rival those obtained on ImageNet when transferring to popular downstream image and video tasks. This differs from previous state-of-the-art approaches to learning image encoders from video, which also obtain such results but require large video datasets, following closely the ImageNet blueprint. 

Our proposed learning method \ours is inspired by DINO, generalizing it to video by incorporating implicit multi-object tracking across video clips.
We observe that the method leads to interesting emergent attention masks within the transformer model, that seem to latch on to particular objects, even through occlusions. This makes it uniquely suited to our newly introduced dataset.

\section{Acknowledgements}
This work was in part supported by the ANR-19-CE23-
0028 MEERQAT project and was performed using the HPC
resources from GENCI-IDRIS Grant 2021 AD011012528. We also thank Dilara Gokay and Andrew Zisserman for their valuable feedback on this paper.

{\small
\bibliographystyle{iclr2024_conference}
\bibliography{refs}

\begin{thebibliography}{68}
\providecommand{\natexlab}[1]{#1}
\providecommand{\url}[1]{\texttt{#1}}
\expandafter\ifx\csname urlstyle\endcsname\relax
  \providecommand{\doi}[1]{doi: #1}\else
  \providecommand{\doi}{doi: \begingroup \urlstyle{rm}\Url}\fi

\bibitem[Adams(1987)]{adams1987evaluation}
Russell~J Adams.
\newblock An evaluation of color preference in early infancy.
\newblock \emph{Infant Behavior and Development}, 1987.

\bibitem[Agrawal et~al.(2015)Agrawal, Carreira, and Malik]{agrawal2015learning}
Pulkit Agrawal, Joao Carreira, and Jitendra Malik.
\newblock Learning to see by moving.
\newblock In \emph{ICCV}, 2015.

\bibitem[Asano et~al.(2020)Asano, Rupprecht, and Vedaldi]{asano2020self}
Yuki~M. Asano, Christian Rupprecht, and Andrea Vedaldi.
\newblock Self-labelling via simultaneous clustering and representation
  learning.
\newblock In \emph{ICLR}, 2020.

\bibitem[Bain et~al.(2021)Bain, Nagrani, Varol, and Zisserman]{bain2021frozen}
Max Bain, Arsha Nagrani, G{\"u}l Varol, and Andrew Zisserman.
\newblock Frozen in time: A joint video and image encoder for end-to-end
  retrieval.
\newblock In \emph{ICCV}, 2021.

\bibitem[Bardes et~al.(2021)Bardes, Ponce, and LeCun]{bardes2021vicreg}
Adrien Bardes, Jean Ponce, and Yann LeCun.
\newblock Vicreg: Variance-invariance-covariance regularization for
  self-supervised learning.
\newblock \emph{arXiv preprint arXiv:2105.04906}, 2021.

\bibitem[Bomba \& Siqueland(1983)Bomba and Siqueland]{bomba1983nature}
Paul~C Bomba and Einar~R Siqueland.
\newblock The nature and structure of infant form categories.
\newblock \emph{Journal of Experimental Child Psychology}, 1983.

\bibitem[Cai \& Vasconcelos(2019)Cai and Vasconcelos]{cai2019cascade}
Zhaowei Cai and Nuno Vasconcelos.
\newblock Cascade r-cnn: High quality object detection and instance
  segmentation.
\newblock \emph{IEEE TPAMI}, 2019.

\bibitem[Campos et~al.(1978)Campos, Hiatt, Ramsay, Henderson, and
  Svejda]{campos1978emergence}
Joseph~J Campos, Susan Hiatt, Douglas Ramsay, Charlotte Henderson, and Marilyn
  Svejda.
\newblock The emergence of fear on the visual cliff.
\newblock \emph{The development of affect}, 1978.

\bibitem[Caron et~al.(2020)Caron, Misra, Mairal, Goyal, Bojanowski, and
  Joulin]{caron2020unsupervised}
Mathilde Caron, Ishan Misra, Julien Mairal, Priya Goyal, Piotr Bojanowski, and
  Armand Joulin.
\newblock Unsupervised learning of visual features by contrasting cluster
  assignments.
\newblock \emph{NeurIPS}, 2020.

\bibitem[Caron et~al.(2021)Caron, Touvron, Misra, J{\'e}gou, Mairal,
  Bojanowski, and Joulin]{caron2021emerging}
Mathilde Caron, Hugo Touvron, Ishan Misra, Herv{\'e} J{\'e}gou, Julien Mairal,
  Piotr Bojanowski, and Armand Joulin.
\newblock Emerging properties in self-supervised vision transformers.
\newblock In \emph{ICCV}, 2021.

\bibitem[Castellano()]{scenedetect}
Brandon Castellano.
\newblock Pyscenedetect.
\newblock \url{https://github.com/Breakthrough/PySceneDetect}.

\bibitem[Central()]{movie_rom}
World~Movie Central.
\newblock The night we met.
\newblock \url{https://www.youtube.com/watch?v=joIzqAueexA}.

\bibitem[Chakraborty \& Namboodiri(2017)Chakraborty and
  Namboodiri]{Chakraborty2017Learning}
Prabuddha Chakraborty and Vinay~P. Namboodiri.
\newblock Learning to estimate pose by watching videos.
\newblock \emph{arXiv preprint arXiv:1704.04081}, 2017.

\bibitem[Chen et~al.(2020)Chen, Kornblith, Norouzi, and Hinton]{chen2020simple}
Ting Chen, Simon Kornblith, Mohammad Norouzi, and Geoffrey Hinton.
\newblock A simple framework for contrastive learning of visual
  representations.
\newblock In \emph{ICML}, 2020.

\bibitem[Chen et~al.(2023)Chen, Peng, Wang, Lu, and Hu]{chen2023seqtrack}
Xin Chen, Houwen Peng, Dong Wang, Huchuan Lu, and Han Hu.
\newblock Seqtrack: Sequence to sequence learning for visual object tracking.
\newblock In \emph{CVPR}, 2023.

\bibitem[Croitoru et~al.(2017)Croitoru, Bogolin, and
  Leordeanu]{croitoru2017unsupervised}
Ioana Croitoru, Simion-Vlad Bogolin, and Marius Leordeanu.
\newblock Unsupervised learning from video to detect foreground objects in
  single images.
\newblock In \emph{ICCV}, 2017.

\bibitem[Cuturi(2013)]{cuturi2013sinkhorn}
Marco Cuturi.
\newblock Sinkhorn distances: Lightspeed computation of optimal transport.
\newblock \emph{NeurIPS}, 2013.

\bibitem[da~Costa et~al.(2022)da~Costa, Fini, Nabi, Sebe, and
  Ricci]{Guilherme2022solo}
Victor Guilherme~Turrisi da~Costa, Enrico Fini, Moin Nabi, Nicu Sebe, and Elisa
  Ricci.
\newblock solo-learn: A library of self-supervised methods for visual
  representation learning.
\newblock \emph{JMLR}, 2022.

\bibitem[Damen et~al.(2022)Damen, Doughty, Farinella, Furnari, Ma, Kazakos,
  Moltisanti, Munro, Perrett, Price, and Wray]{Damen2022RESCALING}
Dima Damen, Hazel Doughty, Giovanni~Maria Farinella, Antonino Furnari, Jian Ma,
  Evangelos Kazakos, Davide Moltisanti, Jonathan Munro, Toby Perrett, Will
  Price, and Michael Wray.
\newblock Rescaling egocentric vision: Collection, pipeline and challenges for
  epic-kitchens-100.
\newblock \emph{IJCV}, 2022.

\bibitem[de~Haan et~al.(2001)de~Haan, Johnson, Maurer, and
  Perrett]{de2001recognition}
Michelle de~Haan, Mark~H Johnson, Daphne Maurer, and David~I Perrett.
\newblock Recognition of individual faces and average face prototypes by 1-and
  3-month-old infants.
\newblock \emph{Cognitive development}, 2001.

\bibitem[Deng et~al.(2009)Deng, Dong, Socher, Li, Li, and
  Fei-Fei]{deng2009imagenet}
Jia Deng, Wei Dong, Richard Socher, Li-Jia Li, Kai Li, and Li~Fei-Fei.
\newblock Imagenet: A large-scale hierarchical image database.
\newblock In \emph{CVPR}, 2009.

\bibitem[Dosovitskiy et~al.(2020)Dosovitskiy, Beyer, Kolesnikov, Weissenborn,
  Zhai, Unterthiner, Dehghani, Minderer, Heigold, Gelly,
  et~al.]{dosovitskiy2020image}
Alexey Dosovitskiy, Lucas Beyer, Alexander Kolesnikov, Dirk Weissenborn,
  Xiaohua Zhai, Thomas Unterthiner, Mostafa Dehghani, Matthias Minderer, Georg
  Heigold, Sylvain Gelly, et~al.
\newblock An image is worth 16x16 words: Transformers for image recognition at
  scale.
\newblock In \emph{ICLR}, 2020.

\bibitem[Everingham et~al.()Everingham, Van~Gool, Williams, Winn, and
  Zisserman]{pascal-voc-2012}
M.~Everingham, L.~Van~Gool, C.~K.~I. Williams, J.~Winn, and A.~Zisserman.
\newblock The {PASCAL} {V}isual {O}bject {C}lasses {C}hallenge 2012 {(VOC2012)}
  {R}esults.
\newblock
  http://www.pascal-network.org/challenges/VOC/voc2012/workshop/index.html.

\bibitem[Gordon et~al.(2020)Gordon, Ehsani, Fox, and
  Farhadi]{gordon2020watching}
Daniel Gordon, Kiana Ehsani, Dieter Fox, and Ali Farhadi.
\newblock \emph{arXiv}, 2020.

\bibitem[Grauman et~al.(2022)Grauman, Westbury, Byrne, Chavis, Furnari,
  Girdhar, Hamburger, Jiang, Liu, Liu, et~al.]{grauman2022ego4d}
Kristen Grauman, Andrew Westbury, Eugene Byrne, Zachary Chavis, Antonino
  Furnari, Rohit Girdhar, Jackson Hamburger, Hao Jiang, Miao Liu, Xingyu Liu,
  et~al.
\newblock Ego4d: Around the world in 3,000 hours of egocentric video.
\newblock In \emph{CVPR}, 2022.

\bibitem[Gu et~al.(2018)Gu, Sun, Ross, Vondrick, Pantofaru, Li,
  Vijayanarasimhan, Toderici, Ricco, Sukthankar, et~al.]{gu2018ava}
Chunhui Gu, Chen Sun, David~A Ross, Carl Vondrick, Caroline Pantofaru, Yeqing
  Li, Sudheendra Vijayanarasimhan, George Toderici, Susanna Ricco, Rahul
  Sukthankar, et~al.
\newblock Ava: A video dataset of spatio-temporally localized atomic visual
  actions.
\newblock In \emph{CVPR}, 2018.

\bibitem[Huang et~al.(2019)Huang, Zhao, and Huang]{huang2019got}
Lianghua Huang, Xin Zhao, and Kaiqi Huang.
\newblock Got-10k: A large high-diversity benchmark for generic object tracking
  in the wild.
\newblock \emph{IEEE TPAMI}, 2019.

\bibitem[Huang et~al.(2021)Huang, Zhao, and Huang]{Huang2021}
Lianghua Huang, Xin Zhao, and Kaiqi Huang.
\newblock Got-10k: A large high-diversity benchmark for generic object tracking
  in the wild.
\newblock \emph{IEEE TPAMI}, 2021.

\bibitem[Jayaraman \& Grauman(2015)Jayaraman and
  Grauman]{jayaraman2015learning}
Dinesh Jayaraman and Kristen Grauman.
\newblock Learning image representations tied to ego-motion.
\newblock In \emph{ICCV}, 2015.

\bibitem[Jayaraman \& Grauman(2016)Jayaraman and Grauman]{jayaraman2016look}
Dinesh Jayaraman and Kristen Grauman.
\newblock Look-ahead before you leap: end-to-end active recognition by
  forecasting the effect of motion.
\newblock In \emph{ECCV}, 2016.

\bibitem[Kakogeorgiou et~al.(2022)Kakogeorgiou, Gidaris, Psomas, Avrithis,
  Bursuc, Karantzalos, and Komodakis]{kakogeorgiou2022hide}
Ioannis Kakogeorgiou, Spyros Gidaris, Bill Psomas, Yannis Avrithis, Andrei
  Bursuc, Konstantinos Karantzalos, and Nikos Komodakis.
\newblock What to hide from your students: Attention-guided masked image
  modeling.
\newblock In \emph{ECCV}, 2022.

\bibitem[Kay et~al.(2017)Kay, Carreira, Simonyan, Zhang, Hillier,
  Vijayanarasimhan, Viola, Green, Back, Natsev, et~al.]{kay2017kinetics}
Will Kay, Joao Carreira, Karen Simonyan, Brian Zhang, Chloe Hillier, Sudheendra
  Vijayanarasimhan, Fabio Viola, Tim Green, Trevor Back, Paul Natsev, et~al.
\newblock The kinetics human action video dataset.
\newblock \emph{arXiv preprint arXiv:1705.06950}, 2017.

\bibitem[Khan et~al.(2020)Khan, Shao, Ali, and Tumrani]{khan2020content}
Abdullah~Aman Khan, Jie Shao, Waqar Ali, and Saifullah Tumrani.
\newblock Content-aware summarization of broadcast sports videos: an
  audio--visual feature extraction approach.
\newblock \emph{Neural Processing Letters}, 2020.

\bibitem[Krause et~al.(2013)Krause, Stark, Deng, and Fei-Fei]{krause20133d}
Jonathan Krause, Michael Stark, Jia Deng, and Li~Fei-Fei.
\newblock 3d object representations for fine-grained categorization.
\newblock In \emph{ICCVW}, 2013.

\bibitem[Krizhevsky et~al.(2009)Krizhevsky, Hinton,
  et~al.]{krizhevsky2009learning}
Alex Krizhevsky, Geoffrey Hinton, et~al.
\newblock Learning multiple layers of features from tiny images.
\newblock 2009.

\bibitem[Li et~al.(2019)Li, Liu, De~Mello, Wang, Kautz, and Yang]{li2019joint}
Xueting Li, Sifei Liu, Shalini De~Mello, Xiaolong Wang, Jan Kautz, and
  Ming-Hsuan Yang.
\newblock Joint-task self-supervised learning for temporal correspondence.
\newblock \emph{NeurIPS}, 2019.

\bibitem[Lin et~al.(2014)Lin, Maire, Belongie, Hays, Perona, Ramanan,
  Doll{\'a}r, and Zitnick]{lin2014microsoft}
Tsung-Yi Lin, Michael Maire, Serge Belongie, James Hays, Pietro Perona, Deva
  Ramanan, Piotr Doll{\'a}r, and C~Lawrence Zitnick.
\newblock Microsoft coco: Common objects in context.
\newblock In \emph{ECCV}, 2014.

\bibitem[Loshchilov \& Hutter(2019)Loshchilov and
  Hutter]{loshchilov2019decoupled}
Ilya Loshchilov and Frank Hutter.
\newblock Decoupled weight decay regularization.
\newblock In \emph{ICLR}, 2019.

\bibitem[Mahendran et~al.(2018)Mahendran, Thewlis, and Vedaldi]{mahendran1}
Aravindh Mahendran, James Thewlis, and Andrea Vedaldi.
\newblock Cross pixel optical-flow similarity for self-supervised learning.
\newblock In \emph{ACCV}, 2018.

\bibitem[Miech et~al.(2019)Miech, Zhukov, Alayrac, Tapaswi, Laptev, and
  Sivic]{miech2019howto100m}
Antoine Miech, Dimitri Zhukov, Jean-Baptiste Alayrac, Makarand Tapaswi, Ivan
  Laptev, and Josef Sivic.
\newblock Howto100m: Learning a text-video embedding by watching hundred
  million narrated video clips.
\newblock In \emph{ICCV}, 2019.

\bibitem[Misra et~al.(2016)Misra, Zitnick, and Hebert]{misra2016shuffle}
Ishan Misra, C~Lawrence Zitnick, and Martial Hebert.
\newblock Shuffle and learn: unsupervised learning using temporal order
  verification.
\newblock In \emph{ECCV}, 2016.

\bibitem[Nilsback \& Zisserman(2008)Nilsback and
  Zisserman]{nilsback2008automated}
Maria-Elena Nilsback and Andrew Zisserman.
\newblock Automated flower classification over a large number of classes.
\newblock In \emph{ICVGIP}, 2008.

\bibitem[Oquab et~al.(2023)Oquab, Darcet, Moutakanni, Vo, Szafraniec, Khalidov,
  Fernandez, Haziza, Massa, El-Nouby, Howes, Huang, Xu, Sharma, Li, Galuba,
  Rabbat, Assran, Ballas, Synnaeve, Misra, Jegou, Mairal, Labatut, Joulin, and
  Bojanowski]{oquab2023dinov2}
Maxime Oquab, Timothée Darcet, Theo Moutakanni, Huy~V. Vo, Marc Szafraniec,
  Vasil Khalidov, Pierre Fernandez, Daniel Haziza, Francisco Massa, Alaaeldin
  El-Nouby, Russell Howes, Po-Yao Huang, Hu~Xu, Vasu Sharma, Shang-Wen Li,
  Wojciech Galuba, Mike Rabbat, Mido Assran, Nicolas Ballas, Gabriel Synnaeve,
  Ishan Misra, Herve Jegou, Julien Mairal, Patrick Labatut, Armand Joulin, and
  Piotr Bojanowski.
\newblock Dinov2: Learning robust visual features without supervision.
\newblock \emph{arXiv:2304.07193}, 2023.

\bibitem[Orhan et~al.(2020)Orhan, Gupta, and Lake]{orhan2020self}
Emin Orhan, Vaibhav Gupta, and Brenden~M Lake.
\newblock Self-supervised learning through the eyes of a child.
\newblock \emph{NeurIPS}, 2020.

\bibitem[Parthasarathy et~al.(2022)Parthasarathy, Eslami, Carreira, and
  H{\'e}naff]{parthasarathy2022self}
Nikhil Parthasarathy, SM~Eslami, Jo{\~a}o Carreira, and Olivier~J H{\'e}naff.
\newblock Self-supervised video pretraining yields strong image
  representations.
\newblock \emph{arXiv preprint arXiv:2210.06433}, 2022.

\bibitem[Pathak et~al.(2017)Pathak, Girshick, Doll{\'a}r, Darrell, and
  Hariharan]{pathak2017learning}
Deepak Pathak, Ross Girshick, Piotr Doll{\'a}r, Trevor Darrell, and Bharath
  Hariharan.
\newblock Learning features by watching objects move.
\newblock In \emph{CVPR}, 2017.

\bibitem[Pirk et~al.(2020)Pirk, Khansari, Bai, Lynch, and
  Sermanet]{pirk2020online}
S{\"o}ren Pirk, Mohi Khansari, Yunfei Bai, Corey Lynch, and Pierre Sermanet.
\newblock Online learning of object representations by appearance space feature
  alignment.
\newblock In \emph{ICRA}, 2020.

\bibitem[Pont-Tuset et~al.(2017)Pont-Tuset, Perazzi, Caelles, Arbel{\'a}ez,
  Sorkine-Hornung, and Van~Gool]{pont20172017}
Jordi Pont-Tuset, Federico Perazzi, Sergi Caelles, Pablo Arbel{\'a}ez, Alex
  Sorkine-Hornung, and Luc Van~Gool.
\newblock The 2017 davis challenge on video object segmentation.
\newblock \emph{arXiv preprint arXiv:1704.00675}, 2017.

\bibitem[Quinn et~al.(1993)Quinn, Eimas, and Rosenkrantz]{quinn1993evidence}
Paul~C Quinn, Peter~D Eimas, and Stacey~L Rosenkrantz.
\newblock Evidence for representations of perceptually similar natural
  categories by 3-month-old and 4-month-old infants.
\newblock \emph{Perception}, 1993.

\bibitem[Ragusa et~al.(2023)Ragusa, Furnari, and
  Farinella]{ragusa_MECCANO_2023}
Francesco Ragusa, Antonino Furnari, and Giovanni~Maria Farinella.
\newblock Meccano: A multimodal egocentric dataset for humans behavior
  understanding in the industrial-like domain.
\newblock \emph{CVIU}, 2023.

\bibitem[Salehi et~al.(2023)Salehi, Gavves, Snoek, and Asano]{salehi2023time}
Mohammadreza Salehi, Efstratios Gavves, Cees G.~M. Snoek, and Yuki~M. Asano.
\newblock Time does tell: Self-supervised time-tuning of dense image
  representations.
\newblock \emph{ICCV}, 2023.

\bibitem[Sener et~al.(2022)Sener, Chatterjee, Shelepov, He, Singhania, Wang,
  and Yao]{sener2022assembly101}
F.~Sener, D.~Chatterjee, D.~Shelepov, K.~He, D.~Singhania, R.~Wang, and A.~Yao.
\newblock Assembly101: A large-scale multi-view video dataset for understanding
  procedural activities.
\newblock \emph{CVPR}, 2022.

\bibitem[Sermanet et~al.(2018)Sermanet, Lynch, Chebotar, Hsu, Jang, Schaal,
  Levine, and Brain]{sermanet2018time}
Pierre Sermanet, Corey Lynch, Yevgen Chebotar, Jasmine Hsu, Eric Jang, Stefan
  Schaal, Sergey Levine, and Google Brain.
\newblock Time-contrastive networks: Self-supervised learning from video.
\newblock In \emph{ICRA}, 2018.

\bibitem[Sim{\'e}oni et~al.(2021)Sim{\'e}oni, Puy, Vo, Roburin, Gidaris,
  Bursuc, P{\'e}rez, Marlet, and Ponce]{simeoni2021localizing}
Oriane Sim{\'e}oni, Gilles Puy, Huy~V Vo, Simon Roburin, Spyros Gidaris, Andrei
  Bursuc, Patrick P{\'e}rez, Renaud Marlet, and Jean Ponce.
\newblock Localizing objects with self-supervised transformers and no labels.
\newblock In \emph{BMVC}, 2021.

\bibitem[Skiptrace()]{movie_action}
Skiptrace.
\newblock Skiptrace.
\newblock \url{https://www.youtube.com/watch?v=LbRNBQaO5a0}.

\bibitem[Sokol(1978)]{sokol1978measurement}
Samuel Sokol.
\newblock Measurement of infant visual acuity from pattern reversal evoked
  potentials.
\newblock \emph{Vision research}, 18\penalty0 (1):\penalty0 33--39, 1978.

\bibitem[Spelke \& Kinzler(2007)Spelke and Kinzler]{spelke2007core}
Elizabeth~S Spelke and Katherine~D Kinzler.
\newblock Core knowledge.
\newblock \emph{Developmental science}, 2007.

\bibitem[Tschannen et~al.(2020)Tschannen, Djolonga, Ritter, Mahendran, Houlsby,
  Gelly, and Lucic]{tschannen2020self}
Michael Tschannen, Josip Djolonga, Marvin Ritter, Aravindh Mahendran, Neil
  Houlsby, Sylvain Gelly, and Mario Lucic.
\newblock Self-supervised learning of video-induced visual invariances.
\newblock In \emph{CVPR}, pp.\  13806--13815, 2020.

\bibitem[Van~Horn et~al.(2018)Van~Horn, Mac~Aodha, Song, Cui, Sun, Shepard,
  Adam, Perona, and Belongie]{van2018inaturalist}
Grant Van~Horn, Oisin Mac~Aodha, Yang Song, Yin Cui, Chen Sun, Alex Shepard,
  Hartwig Adam, Pietro Perona, and Serge Belongie.
\newblock The inaturalist species classification and detection dataset.
\newblock In \emph{CVPR}, 2018.

\bibitem[Wang \& Gupta(2015)Wang and Gupta]{wang2015unsupervised}
Xiaolong Wang and Abhinav Gupta.
\newblock Unsupervised learning of visual representations using videos.
\newblock In \emph{ICCV}, 2015.

\bibitem[Wang et~al.(2019)Wang, Jabri, and Efros]{wang2019learning}
Xiaolong Wang, Allan Jabri, and Alexei~A Efros.
\newblock Learning correspondence from the cycle-consistency of time.
\newblock In \emph{CVPR}, 2019.

\bibitem[Wiles et~al.(2022)Wiles, Carreira, Barr, Zisserman, and
  Malinowski]{wiles2022compressed}
Olivia Wiles, Joao Carreira, Iain Barr, Andrew Zisserman, and Mateusz
  Malinowski.
\newblock Compressed vision for efficient video understanding.
\newblock In \emph{ACCV}, 2022.

\bibitem[Wiskott \& Sejnowski(2002)Wiskott and Sejnowski]{wiskott2002slow}
Laurenz Wiskott and Terrence~J Sejnowski.
\newblock Slow feature analysis: Unsupervised learning of invariances.
\newblock \emph{Neural computation}, 2002.

\bibitem[Xiao et~al.(2018)Xiao, Liu, Zhou, Jiang, and Sun]{xiao2018unified}
Tete Xiao, Yingcheng Liu, Bolei Zhou, Yuning Jiang, and Jian Sun.
\newblock Unified perceptual parsing for scene understanding.
\newblock In \emph{ECCV}, 2018.

\bibitem[Xiong et~al.(2021)Xiong, Ren, Zeng, and Urtasun]{xiong2021self}
Yuwen Xiong, Mengye Ren, Wenyuan Zeng, and Raquel Urtasun.
\newblock Self-supervised representation learning from flow equivariance.
\newblock In \emph{ICCV}, 2021.

\bibitem[Zhou et~al.(2017)Zhou, Zhao, Puig, Fidler, Barriuso, and
  Torralba]{zhou2017scene}
Bolei Zhou, Hang Zhao, Xavier Puig, Sanja Fidler, Adela Barriuso, and Antonio
  Torralba.
\newblock Scene parsing through ade20k dataset.
\newblock In \emph{CVPR}, 2017.

\bibitem[Zhou et~al.(2022{\natexlab{a}})Zhou, Wei, Wang, Shen, Xie, Yuille, and
  Kong]{zhou2021ibot}
Jinghao Zhou, Chen Wei, Huiyu Wang, Wei Shen, Cihang Xie, Alan Yuille, and Tao
  Kong.
\newblock ibot: Image bert pre-training with online tokenizer.
\newblock \emph{ICLR}, 2022{\natexlab{a}}.

\bibitem[Zhou et~al.(2022{\natexlab{b}})Zhou, Girdhar, Joulin,
  Kr{\"a}henb{\"u}hl, and Misra]{zhou2022detecting}
Xingyi Zhou, Rohit Girdhar, Armand Joulin, Philipp Kr{\"a}henb{\"u}hl, and
  Ishan Misra.
\newblock Detecting twenty-thousand classes using image-level supervision.
\newblock In \emph{ECCV}, 2022{\natexlab{b}}.

\end{thebibliography}
}

\clearpage
\appendix
\addcontentsline{toc}{section}{Appendices}
\part{\LARGE{{Is ImageNet worth 1 video? Learning strong image encoders from 1 long unlabelled video}} \\
\vspace{2em}
\Large{Supplementary Material}}

{
\hypersetup{linkcolor=black}
\parttoc 
}

\bigskip

\pagenumbering{arabic}

\renewcommand{\footnotesize}{\fontsize{7pt}{11pt}\selectfont}

\section{More on Walking Tours}
\label{app:wt}

\subsection{Comparison with other video datasets}
\label{app:data-comp}

In \autoref{tab:sota-dataset}, we compare \WT with different types of existing video datasets. Self-supervised pretraining on videos has been mostly limited to 
video datasets that rely on weak annotation in the form of video-text pairs~\citep{bain2021frozen, miech2019howto100m} or even are curated, \eg their class balance is controlled, even if their annotation is unused~\citep{kay2017kinetics}. Their average clip duration is small, \eg less than 20 sec, and their resolution is also small, limiting the capacity to detect objects at a greater distance. By contrast, \WT videos are continuous, hours-long at high resolution and provide natural transitions of scenes and viewing conditions. They are not curated and thus better suited for the self-supervised setting.

\emph{ImageNet-aligned} datasets such as R2V2~\citep{gordon2020watching} and VideoNet~\citep{parthasarathy2022self} contain videos that are curated and annotated with the same distribution and classes as ImageNet, meant for pretraining image encoders. These videos are short, \ie 10 seconds on average. By contrast, \WT consists of a continuous stream of egocentric video, where the average number of classes is close to that of ImageNet, as shown in \autoref{sec:data-analysis}. The rich information contained in 4K resolution, together with a high number of objects in a frame, makes it appropriate for representation learning. Importantly, the continuity and absence of curation make it more realistic and more comparable with human learning. 
Our dataset does not rely on a set of objects, human activities or other search terms but instead is \emph{data-first} and more open-ended.

Despite the large number of high-quality videos, \emph{egocentric} video datasets~\citep{Damen2022RESCALING, grauman2022ego4d, sener2022assembly101} have been used only for downstream tasks and thus come with extensive annotation. In comparison, \WT has 4-10 times longer average duration and twice the frame resolution. While \WT is smaller in terms of total duration and number of videos, it is scalable under the self-supervised setting since it requires no human labeling effort and more videos can be easily found, downloaded or even made. This makes collecting more data as simple as a walk in the park.

\emph{Very long video datasets.} A large dataset of 10k \WT videos was created recently by~\citep{wiles2022compressed} but was not publicly released and not studied for self-supervised learning. Another dataset having hour-long videos is introduced in~\citep{khan2020content}, in the context of sports analytics; it has not been explored for self-supervised learning either.

\subsection{Dataset analysis}
\label{app:data-analysis}

Here we present a more detailed discussion on the dataset analysis of \autoref{sec:data-analysis}. We refer to \autoref{fig:dataset-stats}, where we analyse the properties of a single \WT video compared with videos of the same length from Epic-Kitchens~\citep{Damen2022RESCALING} and AVA~\citep{gu2018ava} datasets, as well as two movie videos, an action movie and a romantic movie.

\paragraph{Variation in lightness}

We measure the change in perceived brightness using the lightness value (L) across consecutive frames. From~\autoref{fig:dataset-stats}(a), we observe a gradual shift at roughly 150 min into the \WT video, transitioning from bright to dim to dark. By contrast, Epic-Kitchens and AVA videos exhibit random brightness fluctuations, alternating between dim and bright conditions. Typically, self-supervised pretraining happens on datasets with uniform brightness levels. Datasets featuring such brightness variations are less expored.

\paragraph{Variation in number of objects}

Using Detic~\citep{zhou2022detecting}, a DETR-style object detector trained on ImageNet-21k, we detect objects in each frame. \autoref{fig:dataset-stats}(b) shows the number of unique objects per frame and \autoref{fig:dataset-stats}(c) shows their frequency in the entire video. We observe that \WT contains 703 unique objects, while Epic-Kitchens has 373,  AVA has 663 and Movie-2 has 259. The unique objects appear more frequently and there are more unique objects per frame in \WT than in the other datasets. This makes \WT semantically richer, despite coming from one continuous stream of video. Using videos with a large number of objects can encourage the model to capture complex relations and variations in the data.

\paragraph{Variation in shots}

Egocentric videos are typically captured in a single uninterrupted take, with exceptions being post-processed special effects or cuts. In~\autoref{fig:dataset-stats}(d), we find that, on average, \WT and Epic-Kitchens videos contain only one or two shots per entire video, while AVA contains 406, an action movie (Movie$_\text{act}$)~\citep{movie_action} contains 2000 and a romantic movie (Movie$_\text{rom}$)~\citep{movie_rom} contains 667. The substantial number of shots in movies and AVA poses challenges for representation learning methods that rely on object tracking or optical flow. In~\autoref{sec:ablation}, we show that \WT significantly outperforms movies in downstream tasks, which may be attributed to the absence of cuts.

\section{More on experimental setup}
\label{app:setup}

\subsection{Multi-crop}
\label{app:multi-crop}

Following DINO~\citep{caron2021emerging} and iBOT~\citep{zhou2021ibot}, we apply the \emph{multi-crop} strategy~\citep{caron2020unsupervised}. In particular, we generate $m$ \emph{local crops} $\vX_t^{\ell_i}$ of smaller resolution for $i \in \{1,\dots,m\}$. The \emph{local loss} $L_t^{\text{LC}}$ for frame $t$ is applied to the \cls token between the teacher $f_{\theta'}$ output for a global view $\vX_t^u$ and the student $f_\theta$ output for the local crop $\vX_t^{\ell_i}$ for $i \in \{1,\dots,m\}$:
\begin{equation}
    L_t^{\text{LC}} \defn \sum_{v \in V} \sum_{i=1}^m
		f_{\theta'} (\vX_t^v)^\cls
		\log \left(f_\theta (\vX_t^{\ell_i})^\cls \right)
\label{eq:local-loss}
\end{equation}
The overall loss $L$ is the sum of the multi-object loss $L_t^{\text{O}}$~\eq{obj-loss} and the local loss $L_t^{\text{LC}}$~\eq{local-loss}, averaged over all $T$ frames:
\begin{equation}
	L \defn \frac{1}{T} \sum_{t=1}^{T} (L_t^{\text{O}} + L_t^{\text{LC}}).
\label{eq:loss}
\end{equation}

\subsection{Implementation details}
\label{app:impl}

We use ViT-S/16~\citep{dosovitskiy2020image} as the backbone in all our experiments. For each mini-batch, we randomly sample clips from the video, consisting of $T=8$ frames temporally separated by 1 second \ie we sample one frame every 30. Objects discovered in the first frame are tracked over the following $7$ frames. Since each frame contains several different objects, applying the standard multi-crop augmentation~\citep{caron2020unsupervised} to the entire frame would result in crops with very different visual content or \emph{noisy} positive pairs. Instead, we apply multi-crop to a $300 \times 300$ crop that we first take from the frame. Following DINO~\citep{caron2021emerging}, we obtain two \emph{global crops} and six \emph{local crops}. Masking~\eq{mask} is applied to the global crops seen by the student for the multi-object loss~\eq{obj-loss}, while local crops are seen directly by the student for the local loss~\eq{local-loss}. We train for 100 epochs by default.

Objects are discovered using attention heads, where the total number of heads is in ViT-S/16 is limited to $h=6$. For the purpose of the ablation of the number $k$ of objects for $k>h$ in \autoref{tab:abl-obj}, we modify the MSA block in the final layer, resulting in configurations of 16 and 32 heads. Consequently, we can identify and track up to 16 and 32 objects within the video clip. To accomplish this, we decompose the query and key embeddings of dimension $d=768$ into 16 and 32 subvectors, resulting in new feature dimensions of 24 and 12 respectively, as opposed to 64 for 6 heads. In~\autoref{tab:abl-obj}, we observe that tracking 16 or 32 objects results in overall poor performance possibly due to the small feature dimension, which encodes poor representations.

\subsection{Hyperparameters}
\label{app:hyper}

\paragraph{ImageNet-1k: Linear probing and $k$-NN}

We pretrain \ours in a self-supervised setting with ViT-S/16 using DINO for 100 and 300 epochs. We use two global and six local crops for each clip and train on 8 A100 GPUs with a global batch size of $16 \times 8 = 128$. We use LARS~\cite{} with a learning rate of $5 \times 10^{-4}$, minimum learning rate of $1 \times 10^{-6}$, global crop scale of $[0.4,1.0]$ and local crop scale $[0.05, 0.4]$.

For linear probing, we follow~\citep{caron2021emerging} and use the frozen features of the transformer backbone to train a linear classifier in a supervised setting. We use global batch size of $1024$ on the training set and evaluate on the validation set of ImageNet-1k. We use top-1 accuracy (\%) as our evaluation metric. For $k$-NN, we freeze the backbone and extract features of training images, then use a $k$-nearest neighbour classifier with $k=20$. 

\paragraph{Pascal-VOC 2012: Object discovery}

We use the validation set of Pascal VOC 2012~\citep{pascal-voc-2012}, which comprises a total of 1449 images. Following LOST~\citep{simeoni2021localizing}, we use the averaged self-attention map, extracted from the final layer of a our pretrained ViT-S/16, to retain $80\%$ of the mass. We use the Jaccard similarity $J$ measured as overlap between predicted mask $P$ and the ground truth mask $G$ as $J (P, G) = \frac{G \cap P}{G \cup P}$. We also use CorLoc, which measures the number of correct predicted boxes, where a predicted box is said to be correct if its IoU $\geq 0.5$.

\paragraph{ADE20k: Semantic segmentation}

We evaluate \ours on ADE20k~\citep{zhou2017scene} for semantic segmentation. The dataset includes 20,000 images in the training set and 2,000 images in the validation set. We use UperNet~\citep{xiao2018unified} as the segmentation model and use \ours pretrained on WT to initialize the backbone. Following the experimental settings in iBOT~\citep{zhou2021ibot}, we use AdamW~\citep{loshchilov2019decoupled} with an initial learning rate of $6 \times 10^{-5}$, weight decay of $1 \times 10^{-2}$, and linear warmup of 1,500 iterations. We fine-tune for 160,000 iterations with a batch size of $4$.

\paragraph{MS-COCO: Object detection}

We evaluate \ours for object detection and instance segmentation on MS-COCO. We use Cascade Mask R-CNN~\citep{cai2019cascade}, which produces bounding boxes and instance masks simultaneously on the COCO dataset. We use a multi-scale training strategy, where we resize images to have a shorter side ranging between $480$ and $800$, ensuring that the longer side does not exceed 1,333 pixels. The learning rate is $1 \times 10^{-4}$ and the weight decay is $0.05$. During training, we fine-tune the entire network using a $1\times$ schedule, which involves 12 epochs with learning rate reductions by a factor of $10$ at epochs $9$ and $11$. We explore different layer decay rates, specifically ${0.65, 0.75, 0.8, 0.9}$, with a rate of 1.0 indicating no decay.

To generate hierarchical feature maps, we utilize the features produced by layers 4, 6, 8, and 12 of our network and apply two deconvolutions for layer 4, one deconvolution for layer 6, identity mapping for layer 8, and max-pooling for layer 12. These post-processing steps enable the creation of hierarchical feature representations. It is important to note that we do not employ multi-scale testing in our experiments.

\paragraph{DAVIS-2017: Video object segmentation}

We assess the performance of \ours for video object segmentation on DAVIS 2017 dataset~\citep{pont20172017}, which involves segmenting between 2 to 4 objects within the video frames. We follow DINO~\citep{caron2021emerging} and evaluate on video frames with a resolution of 480p. We apply label propagation on the attention map from our pretrained model and use mean region-based similarity $\mathcal{J}_m$ and mean contour-based accuracy $\mathcal{F}_m$ as our evaluation metrics.

\paragraph{GOT-10k: Object tracking}

We evaluate the object-tracking performance of \ours on the GOT-10k dataset~\citep{huang2019got}. This is a large-scale benchmark for object tracking that contains 563 categories of common moving objects. The training set contains around 10,000 videos and the test set contains 180 videos. Another challenging aspect of this dataset is that the object classes in the training and test set are non-overlapping. We use the SeqTrack~\citep{chen2023seqtrack} codebase to evaluate the performance of different methods on this dataset. In particular, we initialize the encoder weights of SeqTrack with the self-supervised weights and keep them frozen during training. While training, we only update the parameters of the lightweight decoder which consists of 2 transformer blocks. We use all the default hyperparameters. We report mean average overlap (mAO) and success rate (SR) at different thresholds. The mAO measures the class-balanced average overlap between the ground truth and predicted bounding boxes whereas SR indicates the percentage of accurately tracked ground truth bounding boxes where the overlap crosses a certain threshold.

\section{More ablations}
\label{app:ablation}

\begin{table}[tb]
\footnotesize
	
       \centering
	\begin{subtable}[b]{0.4\linewidth}\centering
		{\centering
\scriptsize
\setlength{\tabcolsep}{3pt}
\begin{tabular}{lcc} \toprule
\Th{Video}       & \Th{LP}        & \Th{CorLoc} \\ \midrule
Amsterdam        & \tb{45.4}      & 54.5        \\
Bangkok          & 42.1           & 54.3        \\
Chiang Mai       & 44.9           & 55.5        \\
Istanbul         & 44.5           & 54.6        \\
Kuala Lampur     & 43.9           & 54.1        \\
Singapore        & 42.7           & 54.7        \\ 
Stockholm        & 44.1           & 54.7        \\
Venice           & 44.5           & \tb{56.2}   \\
Wildlife         & 44.0           & 54.9        \\ 
Zurich           & 44.9           & 54.4        \\ \midrule
Mean             & 44.1           & 54.8        \\ \bottomrule
\end{tabular}
\caption{\emph{WT videos}}
\label{tab:1-WT}

}
	\end{subtable}
    \hfill
	\begin{subtable}[b]{0.55\linewidth}\centering
		{\centering
\scriptsize
\setlength{\tabcolsep}{3pt}
\begin{tabular}{llccccc} \toprule
\Th{Method}    & \Th{PT}          &\Th{LP}    & \Th{CorLoc} \\ \midrule  
DINO           & WT               & 33.8      & 51.2        \\ \midrule
\ours          & Movie            & 35.3      & 51.6        \\
\ours          & Movie$^\dagger$  & 39.8      & 54.8        \\ \rowcolor{LightCyan}
\ours          & WT               & \tb{44.5} & \tb{56.2}   \\ \bottomrule
\end{tabular}
\caption{\emph{Cuts}}
\label{tab:abl-cuts}
}
	\end{subtable} 
\caption{\emph{Effect of pretraining video and cuts}. ViT-S/16 pretrained, then frozen. (a) Different \WT video, using \ours. (b) Effect of cuts. $^*$: subset of videos with same total duration as a single \WT video. $^\dagger$: sampling without cuts. LP: top-1 accuracy (\%) of linear probing on the validation set of ImageNet-1k. CorLoc: correct localization on validation set of Pascal-VOC 2012.}
	\label{table:sota-imagenet}
\end{table}

\paragraph{Blurring faces in videos}
We address the privacy concerns that may arise from potential lack of consent during recording by the creator. Given the significant presence of individuals in WTour videos, we employ Deface\footnote{\url{https://github.com/ORB-HD/deface}} to automatically detect and blur faces in WT videos.  Specifically, we blur all faces in WT-Amsterdam and observe that DoRA achieves a top-1 accuracy of 45.5\% on linear probing, compared to 45.4\% without any blur. This demonstrates that face blurring does not impact performance and can effectively mitigate potential privacy and safety issues.

\paragraph{Pretraining \Wt video}

We study the effect of pretraining on different videos of \WT. In \autoref{tab:1-WT}, we observe that the effect is minimal on both image classification and unsupervised object discovery. Notably, the fluctuation in illumination conditions within the \emph{Bangkok} video influences the performance on image classification. It is also interesting to note that, while pretraining on \emph{Amsterdam} is best on image classification, pretraining on \emph{Venice} is best on object discovery. This could be due to the large overlap of objects in these videos with respect to the downstream datasets. However, the consistency of our method across diverse videos indicates that \ours is robust to variations in scenes, number of objects and lighting conditions.

\paragraph{Presence of cuts}

We now analyse the effect of cuts in representation learning. Cuts are defined as instant transitions from one shot to the next, which is frequent in movies. In action movies, a single shot lasts around 4 seconds, while in romance movies, around 12 seconds on average\footnote{\url{https://stephenfollows.com/many-shots-average-movie/}}. To understand the effect of cuts, we compare pretraining on \WT videos and a romance movie. We use PySceneDetect~\citep{scenedetect} to extract the cut timestamps in the movie and we pretrain \ours by sampling clips that do not intersect cuts; cuts naturally do not exist in \Wt videos. In~\autoref{tab:abl-cuts}, we observe that the performance improves significantly in the absence of cuts, as tracking in \ours will fail across a cut.

\begin{table}
\centering
\scriptsize
\setlength{\tabcolsep}{4pt}
\begin{tabular}{llccccccccc} \toprule
\mr{2}{\Th{Method}}                     & \mr{2}{\Th{Pretrain}}     & \mr{2}{\Th{Epochs}}  & \mc{2}{\Th{Object disc.}} & \mc{2}{\Th{Semantic Seg.}} & \mc{2}{\Th{Object Det.}} & \mc{2}{\Th{VOS}} \\ \cmidrule{4-11}
                                        &    &       & \Th{Jacc} &\Th{CorLoc}    & \Th{mIoU} & \Th{Acc$_m$}   &\Th{mAP} & \Th{mIoU}  & \Th{$\cJ_m$} &$\mathcal{F}_m$    \\ \midrule
DINO\tiny~\citep{caron2021emerging}     & IN-1K     & 100     & 44.5  & 59.6       & 33.9  & 44.3           & 39.9    & 35.1     & 57.4    & 61.4      \\
DINO\tiny~\citep{caron2021emerging}     & IN-1K     & 300     & \tb{45.9}  & \tb{64.0}    & \tb{38.2}  & \tb{49.4}     & \tb{42.4}  & \tb{41.6}     & \tb{60.2}   & \tb{63.4}           \\ \midrule

DINO\tiny~\citep{caron2021emerging}     & \Wtv      & 100     & 43.8  & 51.2       & 32.4  & 43.7           & 37.1    & 32.1     & 53.0    & 56.2       \\
\rowcolor{LightCyan}
\ours                                   & \Wtv      & 100     & 44.0  & 56.2       & 35.4  & 45.5           & 39.5    & 34.7     & 56.4    & 60.4      \\
\rowcolor{LightCyan}
\ours                                   & \Wtv      & 300     & \tb{44.9}  & \tb{56.9}       & \tb{37.5}  & \tb{47.3}      & \tb{42.8}    & \tb{42.0}     & \tb{57.5}    & \tb{60.9}     \\ \midrule

DINO\tiny~\citep{caron2021emerging}     & \Wta      & 100     & 42.9  & 55.8       & 34.1  & 44.0           & 38.3    & 33.2     & 53.3        & 57.6     \\
\rowcolor{LightCyan}
\ours                                   & \Wta      & 100     & 44.3  & 57.1       & 36.9  & 49.0           & 40.7    & 36.3     & 55.1    & 60.2     \\
\rowcolor{LightCyan}
\ours                                   & \Wta      & 300     & \tb{45.2}  & \tb{61.3}       & \tb{38.7}  & \tb{49.8}      & \tb{43.7}   & \tb{42.6}     & \tb{58.4}     & \tb{62.0}     \\ \bottomrule
\end{tabular}
\vspace{-3pt}
\caption{\emph{Effect of longer pretraining} ViT-S/16 pretrained, then frozen (object discovery and VOS, same settings as \autoref{tab:sota-cls},~\autoref{tab:sota-cls}) or fine-tuned (semantic segmentation and object detection, same settings as \autoref{tab:ade-coco}). IN-1K: ImageNet-1K.}
\label{tab:long-training}
\end{table}

\paragraph{Longer pretraining}
We investigate the impact of pretraining DoRA on a single WT video and all WT videos over 300 epochs. The results presented in \autoref{tab:long-training} indicate a notable enhancement in performance when DoRA is pretrained for 300 epochs in general. Furthermore, we observe that DoRA pretrained on \Wta surpasses \Wtv in VOS pretraining performance, in contrast to the situation where DoRA was pretrained for 100 epochs. It is worth highlighting that the extended pretraining duration also enables DoRA to achieve comparable results to DINO pretrained on ImageNet-1K, while demonstrating superior performance in dense scene understanding tasks.

\paragraph{Number of frames in a video clip}
We study the effectiveness of DoRA in classification, object discovery and semantic segmentation tasks using a lower number of frames in a video clip, specifically when $T \in \{2,4,6\}$. Using the frame-wise loss~\eq{obj-loss}, our goal is to improve the training efficiency of DoRA by achieving comparable performance through pretraining with fewer frames than the default $T=8$ frames.

\begin{table}
\centering
\setlength{\tabcolsep}{4pt}
\begin{tabular}{lcccccc} \toprule
\mr{2}{\Th{\# Frames}}     & \mc{2}{\Th{Classification}}  & \mc{2}{\Th{Object disc.}} & \mc{2}{\Th{Semantic Seg.}} \\ \cmidrule{2-7}
                           & \Th{LP} &\Th{$k$-NN}         & \Th{Jacc} &\Th{CorLoc}    & \Th{mIoU} & \Th{Acc$_m$}   \\ \midrule
DoRA ($T=2$)               & 43.1    & 32.6               & 41.9      & 53.3          & 32.0       & 43.5            \\
DoRA ($T=4$)               & 44.0    & 32.9               & 42.7      & 54.6          & 32.6       & \tb{44.1}           \\ 
DoRA ($T=6$)               & 44.9    & 33.5               & \tb{44.0} & 55.8          & \tb{32.7}  & 43.9           \\
DoRA ($T=8$)               & \tb{45.4}    & \tb{33.8}     & \tb{44.0} & \tb{56.2}     & 32.4       & 43.7           \\ \bottomrule
\end{tabular}
\vspace{-3pt}
\caption{\emph{Effect of \# frames in a video clip}. ViT-S/16 pretrained on \Wtv, then frozen features used for classification on ImageNet-1K and object discovery on PASCAL-VOC; features fine-tuned on ADE20K for semantic segmentation.}
\label{tab:num-frames}
\end{table}

We observe from~\autoref{tab:finetuning} that the performance of DoRA in classification and object discovery improves as the number of frames in a video clip increases. Employing $T=2$ enhances training throughput by 60\%, albeit with a 2.3\% decrease in linear probing accuracy. However, fine-tuning DoRA for semantic segmentation on ADE20K shows no additional performance gain when varying the number of frames in the video clip. These insights contribute to a detailed understanding of DoRA's performance across diverse tasks and provide valuable guidance in optimizing its training strategy to suit specific applications.

\paragraph{Using larger ViT architecture}

We evaluate DoRA (WT-Venice) for 100 epochs on ViT-B/16 for semantic segmentation on ADE20K and object detection on MS-COCO. The results are summarized in the Table below.

\begin{table}[!h]
\centering
\small
\setlength{\tabcolsep}{3pt}
\begin{tabular}{lcc|cccc} \toprule
	\Th{Method}     & \Th{Arch}    & \Th{Pretrain}  & \Th{ADE20K}  & \Th{MS-COCO}     \\ \midrule
	DoRA (ours)     & ViT-S/16     & WT-Venice      & 35.4           & 39.5           \\
        DoRA (ours)     & ViT-B/16     & WT-Venice      & \textbf{40.3}  & \textbf{41.7}  \\ \bottomrule
\end{tabular}
\caption{\emph{Using a larger model.} ViT-B/16 pretrained and then fine-tuned for semantic segmentation on ADE20K and object detection on MS-COCO. We report mIoU on ADE20K and mAP on MS-COCO.}
\end{table}

We observe that with ViT-B/16, DoRA achieves 4.9\% gain in terms of mIoU on ADE20K and 2.2\% in terms of mAP on MS-COCO as compared to using ViT-S/16. This shows that DoRA also scales well with the model size.

\section{More comparisons with state of the art}
\label{app:results}

\paragraph{Pretraining on ImageNet-1k}

We pretrain \ours on ImageNet-1k and compare with SoTA methods on multiple tasks. Unlike videos, we discover objects but do not track them. Instead, images in a mini-batch are processed independently. Given an input image $\vX$, we obtain refined object prototypes as usual~\eq{proto-2}, but the refined cross-attention~\eq{cross-2} is with $\Tilde{K}_t$ replaced by $\Tilde{K}$ of the same image $\vX$. The same image $\vX$ is masked for the student~\eq{mask}. The loss is given again by~\eq{obj-loss} and~\eq{local-loss} with $\vX_t$ replaced by $\vX$, averaged over the mini-batch. We refer to this version as \emph{\ours without tracking} or \our$^\star$.

\begin{table}
\centering
\scriptsize
\setlength{\tabcolsep}{4pt}
\begin{tabular}{lccccccccc} \toprule
\mr{2}{\Th{Method}}                 & \mr{2}{\Th{Epochs}} & \mc{2}{\Th{Classification}}  & \mc{2}{\Th{Object disc.}} & \mc{2}{\Th{Semantic Seg.}} & \mc{2}{\Th{Object Det.}} \\ \cmidrule{3-10}
                                    &                     & \Th{LP} &\Th{$k$-NN}         & \Th{Jacc} &\Th{CorLoc}    & \Th{mIoU} & \Th{Acc$_m$}   &\Th{mAP} & \Th{mIoU}      \\ \midrule
DINO\tiny~\citep{caron2021emerging} & 100                 & 71.4    & 69.0               & 44.5      & 59.6          & 33.9      & 44.3           & 37.1    & 32.1           \\
iBOT\tiny~\citep{zhou2021ibot}      & 100                 & 72.1    & 69.4               & 44.5      & 59.7          & 35.2      & 45.1           & 38.9    & 34.4           \\ \midrule \rowcolor{LightCyan}
\our$^\star$  (ours)                      & 60                  & 71.9    & 69.4               & 44.4      & 60.0          & 35.4      & 44.9           & 39.3    & 34.9           \\ \rowcolor{LightCyan}
\our$^\star$ (ours)                        & 100                 & 72.2    & 69.6               & 44.8      & 60.2          & 35.8      & 45.1           & 39.9    & 35.1           \\ \bottomrule
\end{tabular}
\vspace{-3pt}
\caption{\emph{Pretraining on ImageNet-1k}. ViT-S/16 pretrained, then frozen (classification and object discovery, same settings as \autoref{tab:sota-cls}) or fine-tuned (semantic segmentation and object detection, same settings as \autoref{tab:ade-coco}). \our$^\star$: \ours without tracking; when pretrained for 60 epochs, it has the same training time as DINO and iBOT.}
\label{tab:sota-imnet}
\end{table}

DINO~\citep{caron2021emerging} and iBOT~\citep{zhou2021ibot} use only one global crop for the student, while \ours uses $k$ object crops. To compensate, we perform an experiment where we pretrain \our$^\star$ for 60 epochs and the competitors for 100, thus all methods having the same training time.

From~\autoref{tab:sota-imnet}, we observe that \ours outperforms state-of-the-art self-supervised learning (SSL) methods like DINO and iBOT on image downstream tasks. This demonstrates that the multi-object loss not only enhances performance when pretrained on \WT videos but also achieves superior results when pretrained on ImageNet-1k images.

\paragraph{Fine-tuning for classification}
We follow the evaluation protocol of iBOT~\citep{zhou2021ibot} and fine-tune DINO (ImNet), DoRA (WT-Venice) and DoRA (WT-all) using ViT-S/16 on CIFAR-10/100 (C-10,C-100)~\citep{krizhevsky2009learning}, iNaturalist18 (iNat 18)~\citep{van2018inaturalist}, Oxford Flowers (Flwrs)~\citep{nilsback2008automated}, Stanford Cars (Cars)~\citep{krause20133d} and ImageNet-1k (ImNet) datasets.

\begin{table}
\centering
\setlength{\tabcolsep}{4pt}
\begin{tabular}{lc|ccccccc} \toprule
	\Th{Method}    & \Th{Pretrained}  & \Th{C-10}  & \Th{C-100}  & \Th{iNat 18}  & \Th{Flwrs}  & \Th{Cars} &\Th{ImNet}  \\ \midrule
        DINO            & IN-1k           & 98.7       & 89.8        & 71.5          &  98.3       & 92.2      & 81.3       \\ \midrule
        \rowcolor{LightCyan}
	DoRA (ours)     & \Wtv       & 98.5       & 89.4        & 69.8          &  94.0       & 92.5      & 80.8           \\
        \rowcolor{LightCyan}
        DoRA (ours)     & \Wta          & \textbf{98.8}       & \textbf{89.9}        & \textbf{72.2}          &  \textbf{98.7}       & \textbf{93.1}      & \textbf{81.4}           \\ \bottomrule
\end{tabular}
\vspace{-3pt}
\caption{\emph{Fine-tuning for classification}. ViT-S/16 pretrained, then fine-tuned on different image-based datasets. \Wtv: Walking Tours (ours), single video of \emph{Venice}; \Wta: all videos. IN-1k: ImageNet-1k. We report the top-1 accuracy (\%).}
\label{tab:finetuning}
\end{table}

From~\autoref{tab:finetuning}, we observe that despite the class distribution of iNat18 being closer to ImNet than WTours, DoRA (WT-Venice) is on-par with DINO (ImNet) when fine-tuned on C-10/C-100, iNat18, Flowers and Cars. Furthermore, DoRA (WT-all) outperforms DINO (ImNet) on all fine-tuning tasks. We also observe such performance gains when fine-tuning on semantic segmentation on ADE20K and object detection on MS-COCO (\autoref{tab:ade-coco}).

\section{More visualizations}
\label{app:viz}

\begin{figure}
\centering
\small
\centering
\setlength{\tabcolsep}{1.5pt}
\newcommand{\sz}{.18}
\newcommand{\cen}{.08\linewidth}
\begin{tabular}{cccccc}
&
$t=1$ &
$t=2$ &
$t=3$ &
$t=4$ &
$t=5$ \\
\raisebox{\cen}{$\vX_t$} &
\fig[\sz]{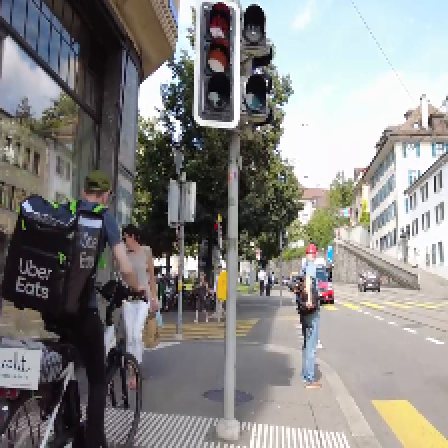} &
\fig[\sz]{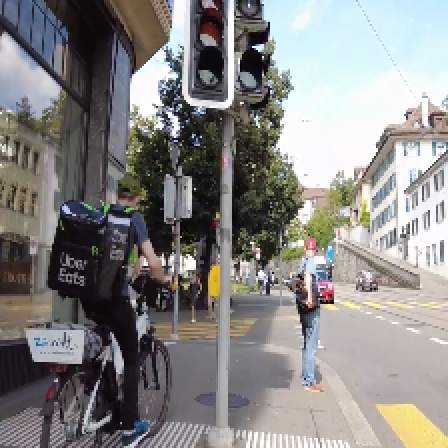} &
\fig[\sz]{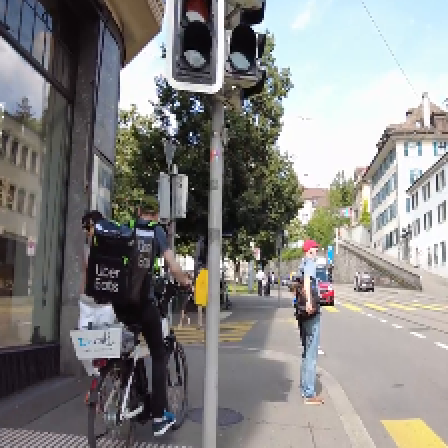} &
\fig[\sz]{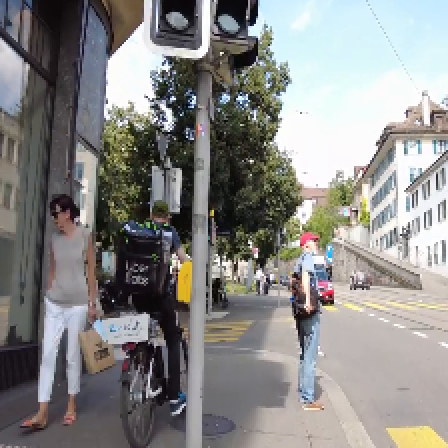} &
\fig[\sz]{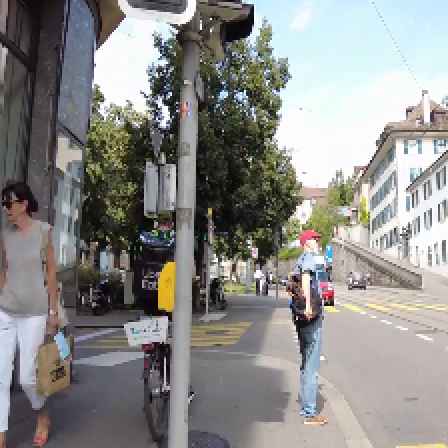} \\
\raisebox{\cen}{$T'_t$} &
\fig[\sz]{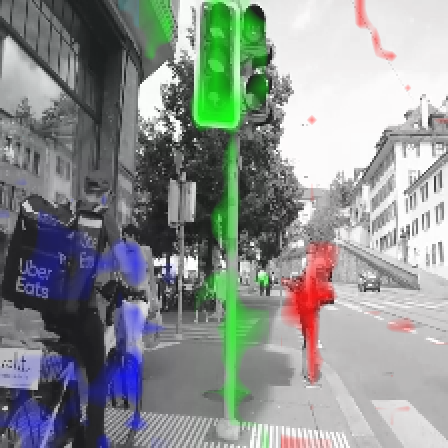} &
\fig[\sz]{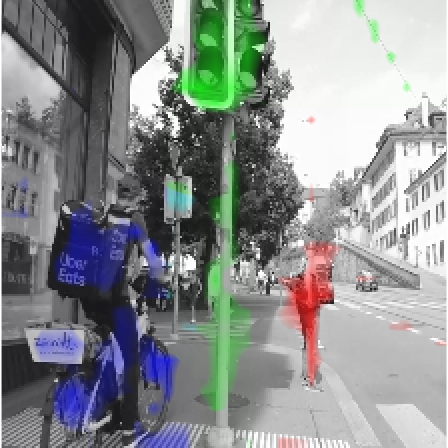} &
\fig[\sz]{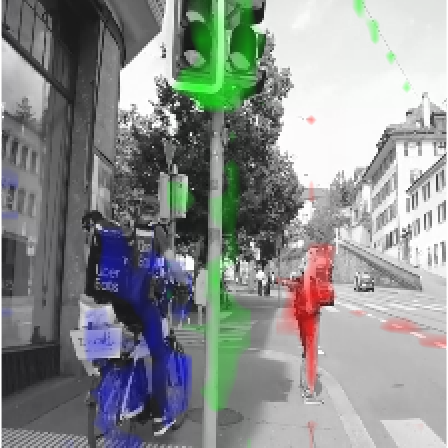} &
\fig[\sz]{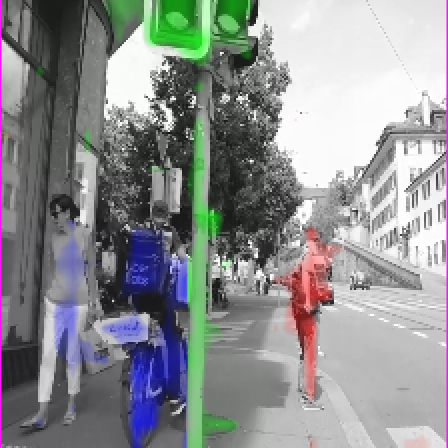} &
\fig[\sz]{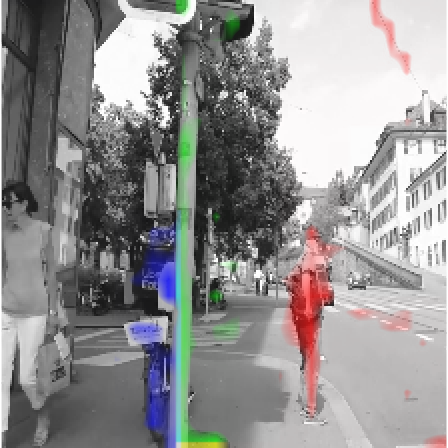} \\
& \mc{5}{(a)} \\[6pt]
\raisebox{\cen}{$\vX_t$} &
\fig[\sz]{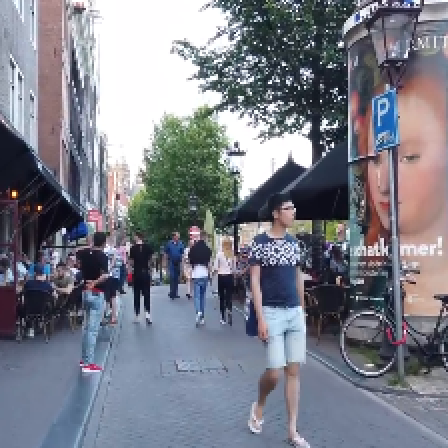} &
\fig[\sz]{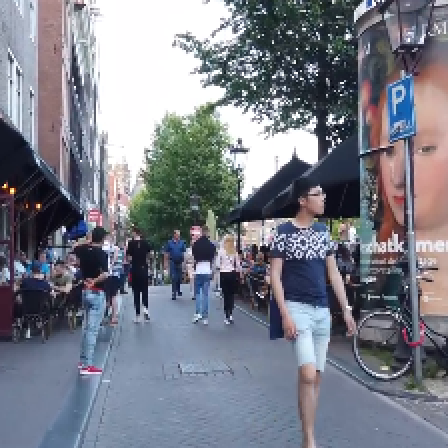} &
\fig[\sz]{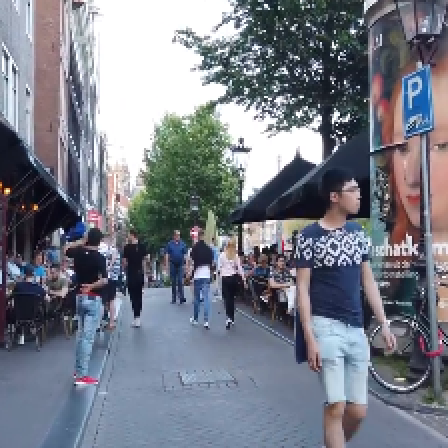} &
\fig[\sz]{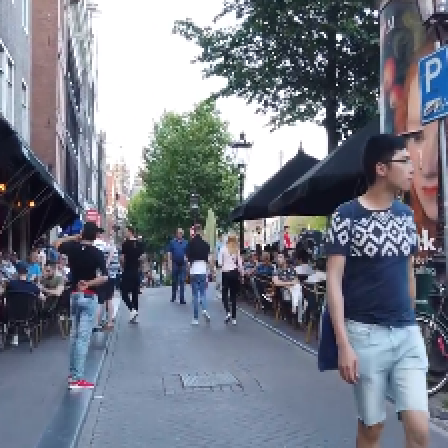} &
\fig[\sz]{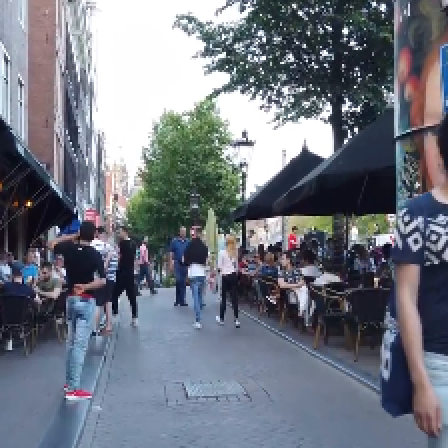} \\
\raisebox{\cen}{$T'_t$} &
\fig[\sz]{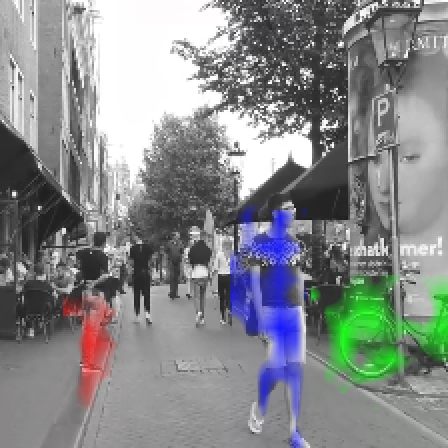} &
\fig[\sz]{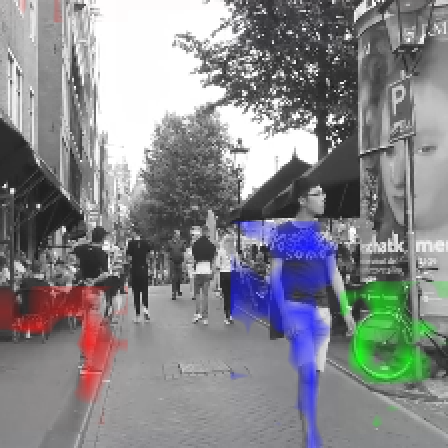} &
\fig[\sz]{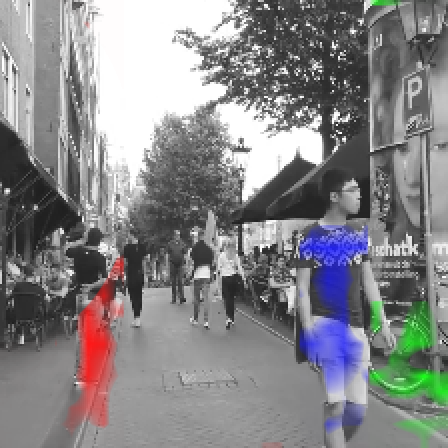} &
\fig[\sz]{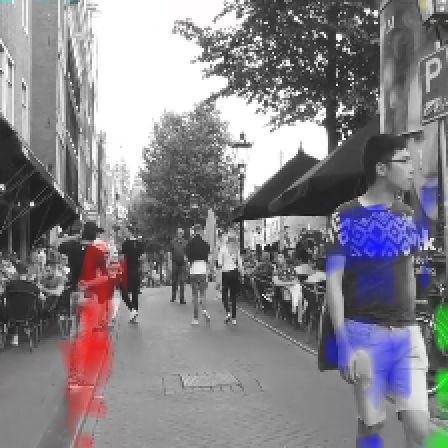} &
\fig[\sz]{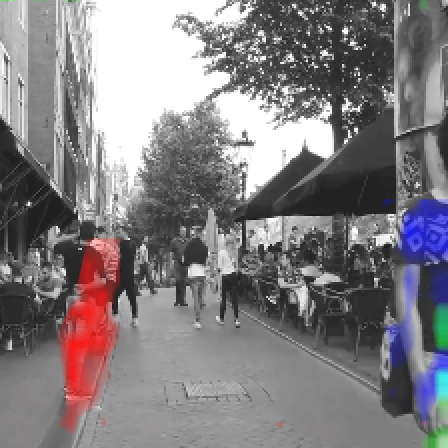} \\
& \mc{5}{(b)} \\[6pt]
\raisebox{\cen}{$\vX_t$} &
\fig[\sz]{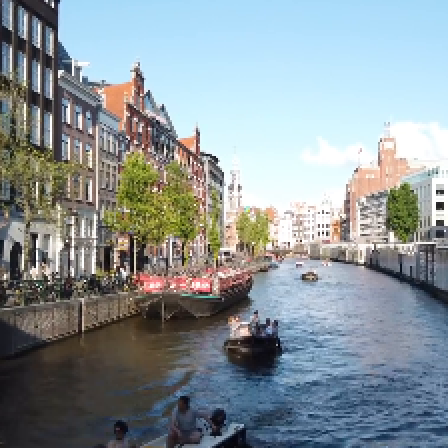} &
\fig[\sz]{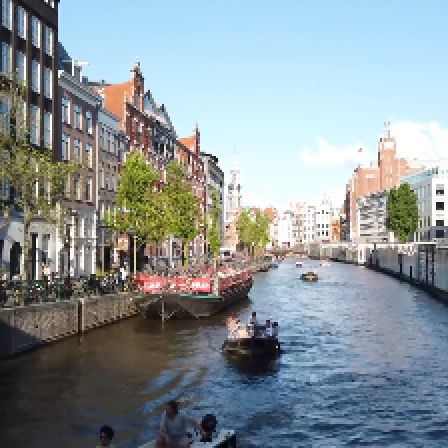} &
\fig[\sz]{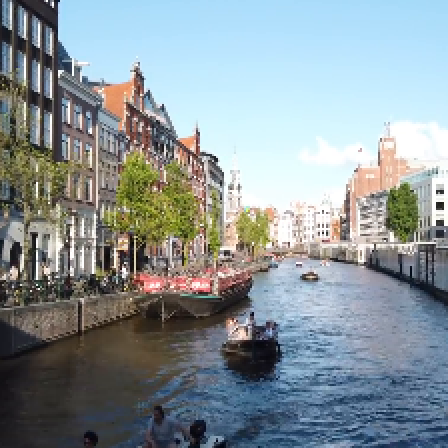} &
\fig[\sz]{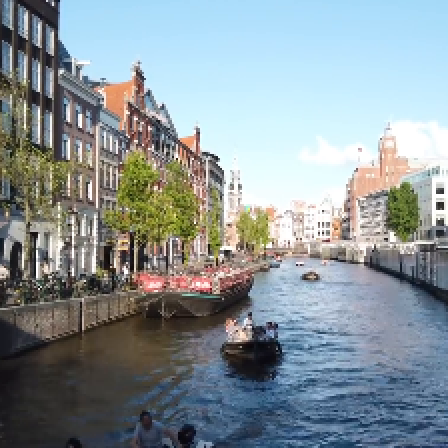} &
\fig[\sz]{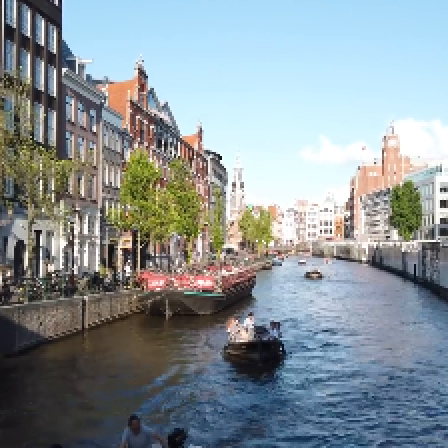} \\
\raisebox{\cen}{$T'_t$} &
\fig[\sz]{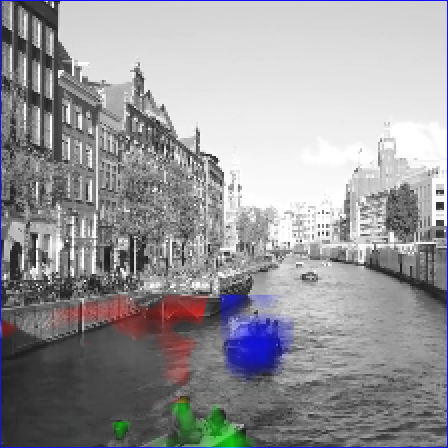} &
\fig[\sz]{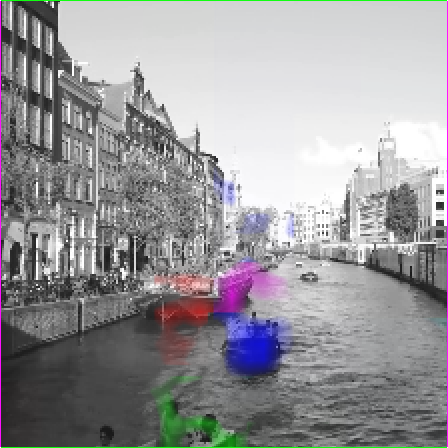} &
\fig[\sz]{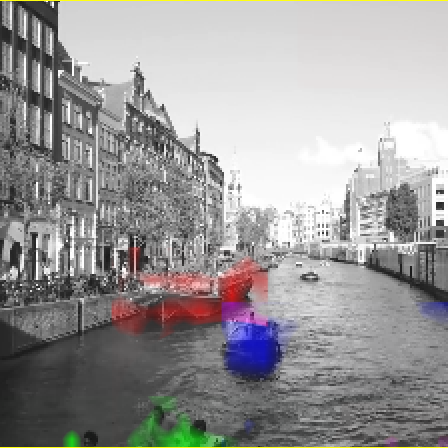} &
\fig[\sz]{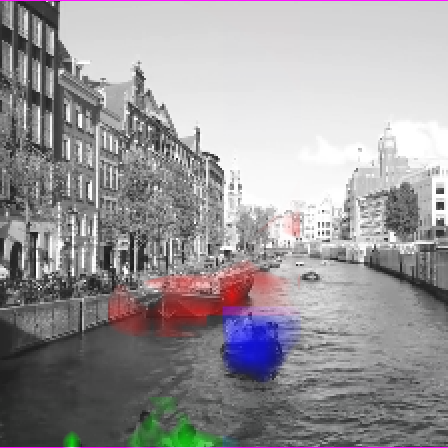} &
\fig[\sz]{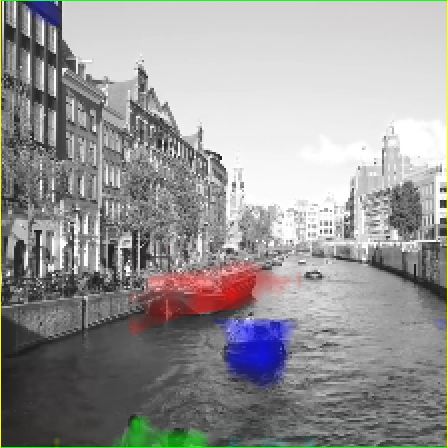} \\
& \mc{5}{(c)} \\
\end{tabular}
\caption{For each input frame $\vX_t$ of a video clip, refined cross-attention map $T'_t \in \real^{k \times n}$~\eq{cross-2}, using Sinkhorn-Knopp. For each object, one row of $T'_t$ is reshaped as $h/p \times w/p$ and upsampled to an $h \times w$ \emph{attention map} overlaid on the input frame for $k=3$ objects encoded in blue, red and green channel. $T'_t$ yields three well separated objects shown in  blue, red and green.}
\label{fig:tracking-app-1}
\end{figure}

\begin{figure}
\centering
\small
\centering
\setlength{\tabcolsep}{1.5pt}
\newcommand{\sz}{.18}
\newcommand{\cen}{.08\linewidth}
\begin{tabular}{cccccc}
&
$t=1$ &
$t=2$ &
$t=3$ &
$t=4$ &
$t=5$ \\
\raisebox{\cen}{$\vX_t$} &
\fig[\sz]{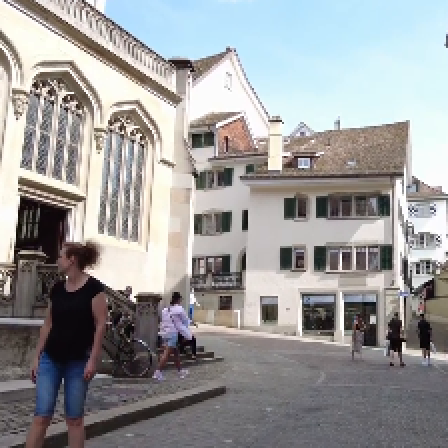} &
\fig[\sz]{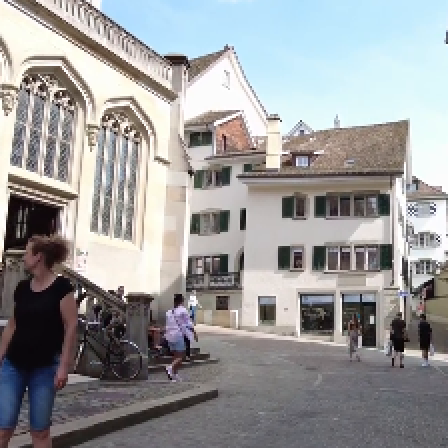} &
\fig[\sz]{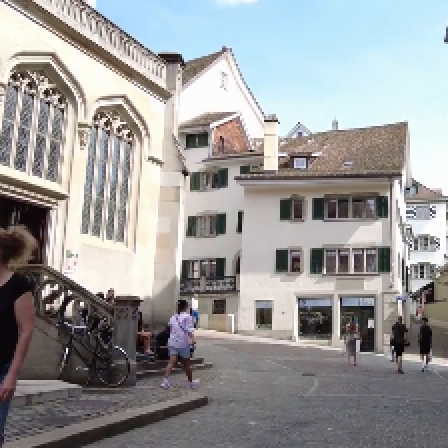} &
\fig[\sz]{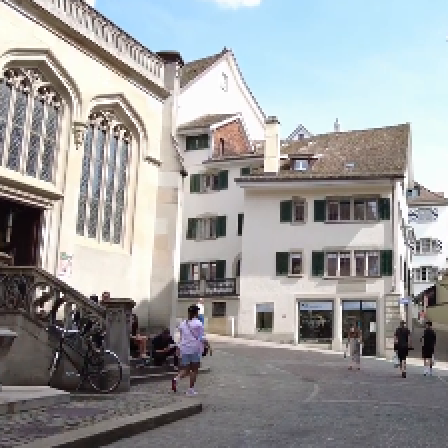} &
\fig[\sz]{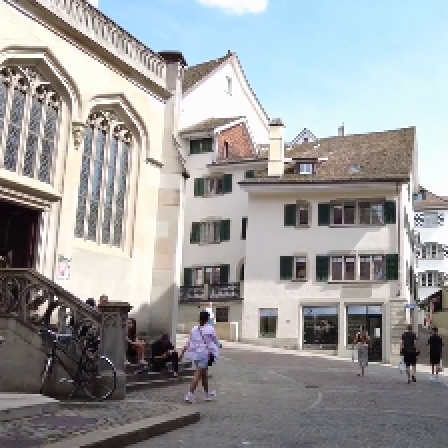} \\
\raisebox{\cen}{$T'_t$} &
\fig[\sz]{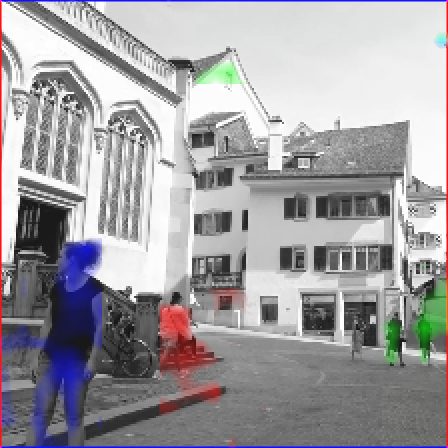} &
\fig[\sz]{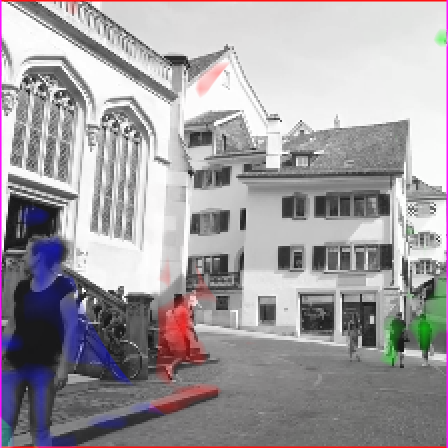} &
\fig[\sz]{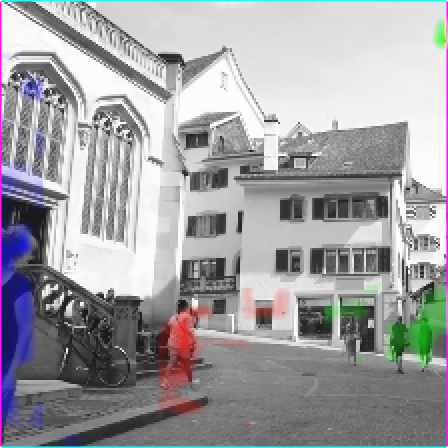} &
\fig[\sz]{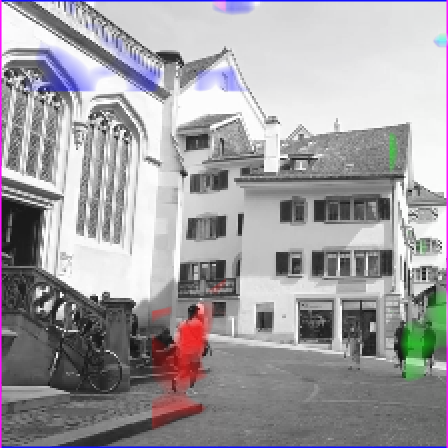} &
\fig[\sz]{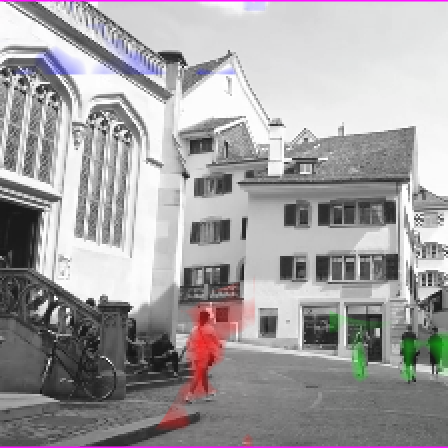} \\
& \mc{5}{(a)} \\[6pt]
\raisebox{\cen}{$\vX_t$} &
\fig[\sz]{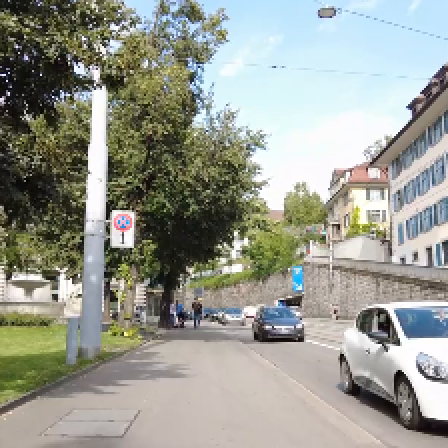} &
\fig[\sz]{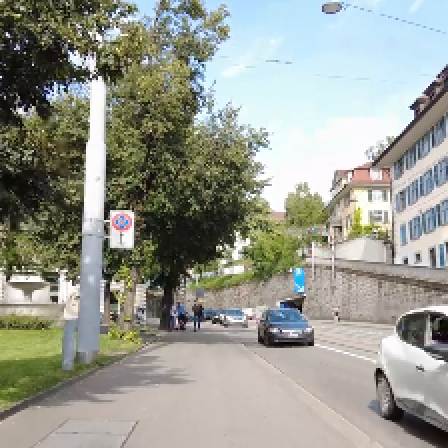} &
\fig[\sz]{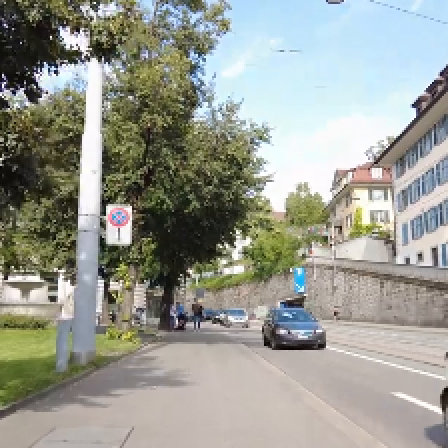} &
\fig[\sz]{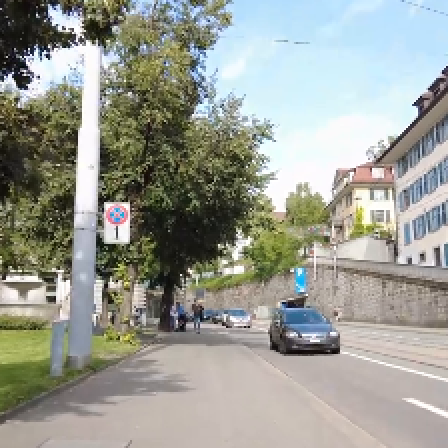} &
\fig[\sz]{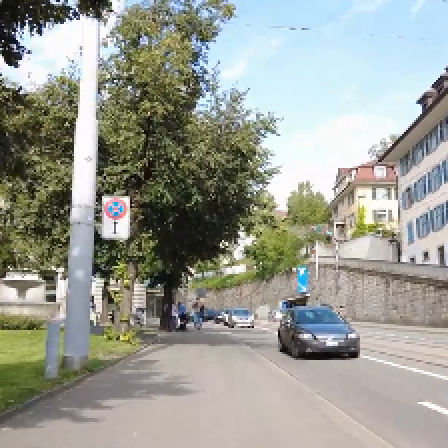} \\
\raisebox{\cen}{$T'_t$} &
\fig[\sz]{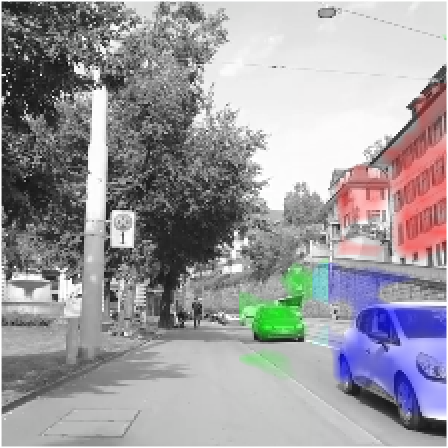} &
\fig[\sz]{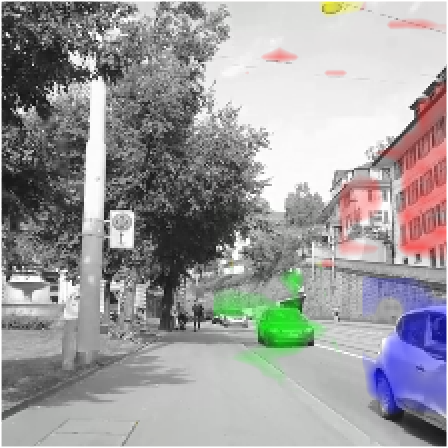} &
\fig[\sz]{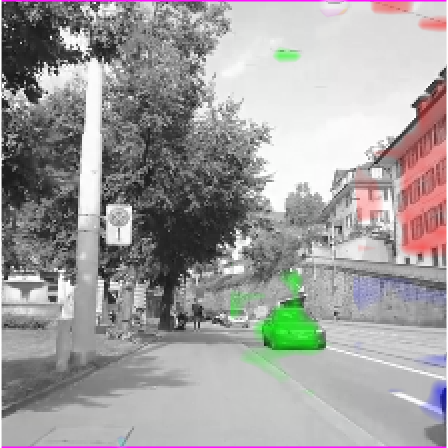} &
\fig[\sz]{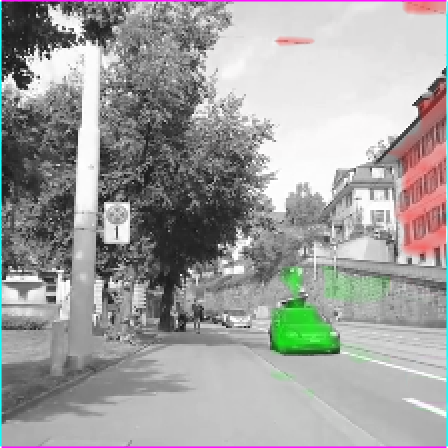} &
\fig[\sz]{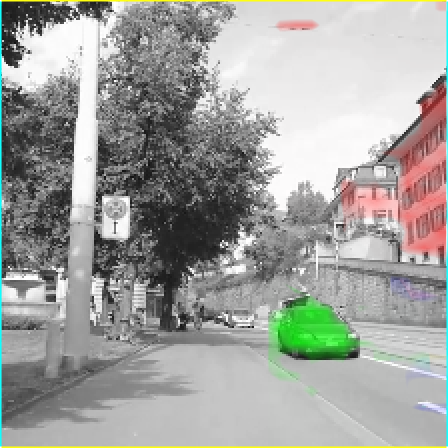} \\
& \mc{5}{(b)} \\[6pt]
\raisebox{\cen}{$\vX_t$} &
\fig[\sz]{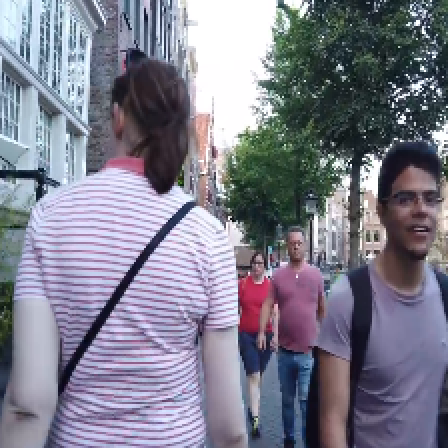} &
\fig[\sz]{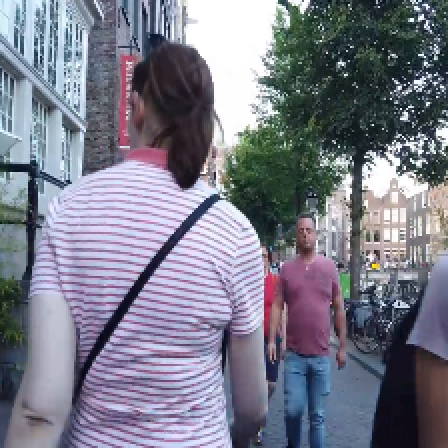} &
\fig[\sz]{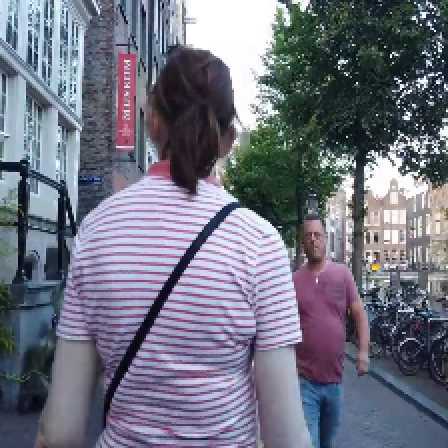} &
\fig[\sz]{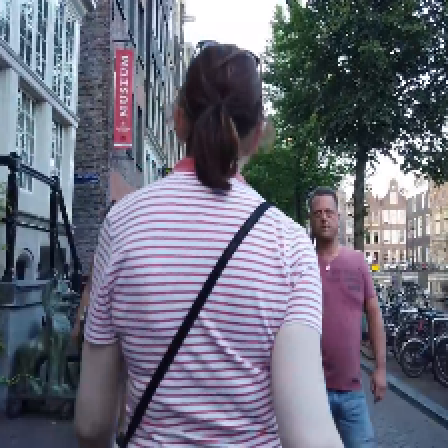} &
\fig[\sz]{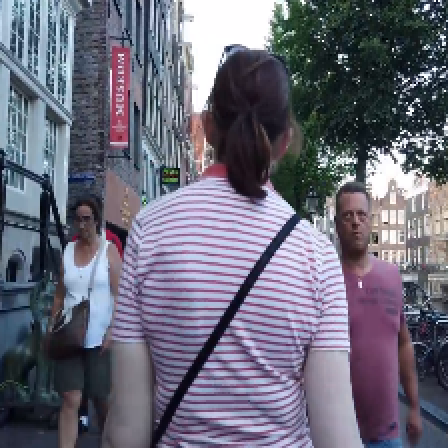} \\
\raisebox{\cen}{$T'_t$} &
\fig[\sz]{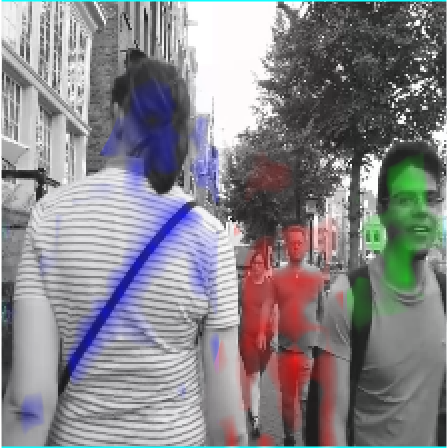} &
\fig[\sz]{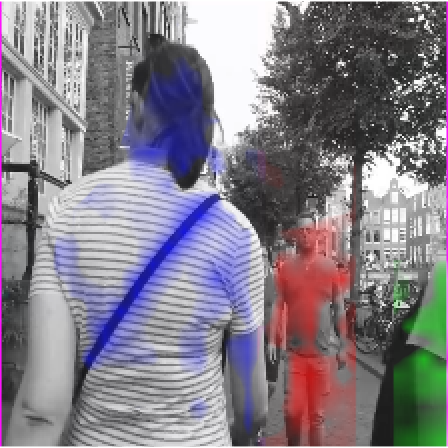} &
\fig[\sz]{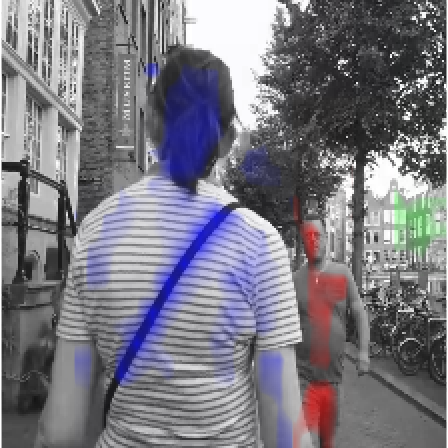} &
\fig[\sz]{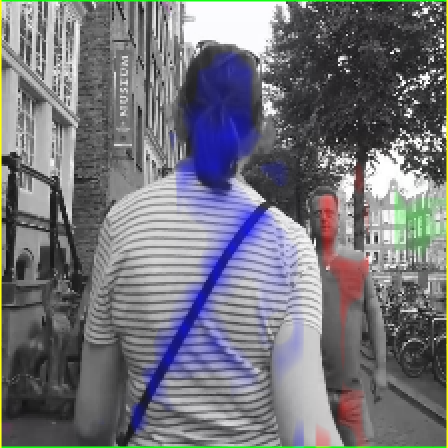} &
\fig[\sz]{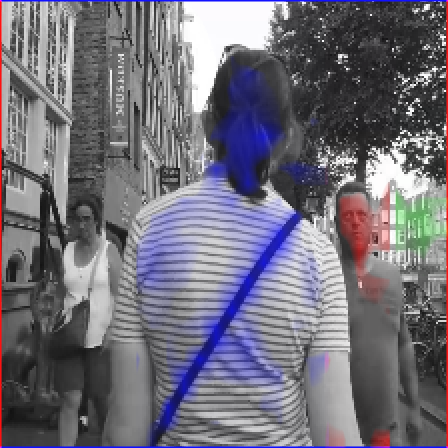} \\
& \mc{5}{(c)} \\
\end{tabular}
\caption{For each input frame $\vX_t$ of a video clip, refined cross-attention map $T'_t \in \real^{k \times n}$~\eq{cross-2}, using Sinkhorn-Knopp. For each object, one row of  $T'_t$ is reshaped as $h/p \times w/p$ and upsampled to an $h \times w$ \emph{attention map} overlaid on the input frame for $k=3$ objects encoded in blue, red and green channel. $T'_t$  yields three well separated objects shown in  blue, red and green.}
\label{fig:tracking-app-2}
\end{figure}

Figures~\ref{fig:tracking-app-1} and~\ref{fig:tracking-app-2} show example attention maps obtained using SK on different clips. These figures show that SK~\eq{cross-2} leads to attention maps that exhibit spatial locality and are well aligned with objects in the scene. Remarkably, the masks seem to be even robust to occlusions, as shown in the sequence with a bicycle moving behind traffic lights.

\end{document}